\documentclass[format=acmsmall, review=false, screen=true]{acmart}

\usepackage{booktabs} 
\usepackage{algorithm}
\usepackage{booktabs} 
\usepackage{algorithmic}
\usepackage{graphicx}
\usepackage{multirow}

\acmJournal{TWEB}
\acmVolume{9}
\acmNumber{4}
\acmArticle{1}
\acmYear{2018}
\acmMonth{4}
\copyrightyear{2018}

\setcopyright{acmlicensed}

\acmDOI{0000001.0000001}{\small \label{key}}

\received{April 2018}
\begin{document}
\title{Exploiting the Value of the Center-dark Channel Prior for Salient Object Detection}

\author{Chunbiao Zhu}
\orcid{0000-0002-6525-6686}  
\affiliation{%
  \institution{SECE, Shenzhen Graduate School, Peking University}
  \streetaddress{2199 Lishui Rd}
  \city{Shenzhen}
  \state{Guangdong}
  \postcode{518055}
  \country{China}
}
\email{zhuchunbiao@pku.edu.cn}

\author{Wenhao Zhang}
\affiliation{%
	\institution{SECE, Shenzhen Graduate School, Peking University}
	\streetaddress{2199 Lishui Rd}
	\city{Shenzhen}
	\state{Guangdong}
	\postcode{518055}
	\country{China}
}
\email{wendy\_zhang@pku.edu.cn}

\author{Thomas H. Li}
\affiliation{%
	\institution{Gpower Semiconductor Inc}
	\city{Suzhou}
	\state{Jiangsu}
	\country{China}
}
\email{thomas.li@gpower-semi.com}

\author{Ge Li}
\affiliation{%
	\institution{SECE, Shenzhen Graduate School, Peking University}
	\streetaddress{2199 Lishui Rd}
	\city{Shenzhen}
	\state{Guangdong}
	\postcode{518055}
	\country{China}
}
\email{geli@ece.pku.edu.cn}

\begin{abstract}
Saliency detection aims to detect the most attractive objects in images and is widely used as a foundation for various applications. In this paper, we propose a novel salient object detection algorithm for RGB-D images using center-dark channel priors. First, we generate an initial saliency map based on a color saliency map and a depth saliency map of a given RGB-D image. Then, we generate a center-dark channel map based on center saliency and dark channel priors. Finally, we fuse the initial saliency map with the center dark channel map to generate the final saliency map. Extensive evaluations over four benchmark datasets demonstrate that our proposed method performs favorably against most of the state-of-the-art approaches. Besides, we further discuss the application of the proposed algorithm in small target detection and demonstrate the universal value of center-dark channel priors in the field of object detection.
\end{abstract}

%
%


\maketitle
\begin{figure}[!h]
	\begin{center}
		\includegraphics[width=0.9\textwidth]{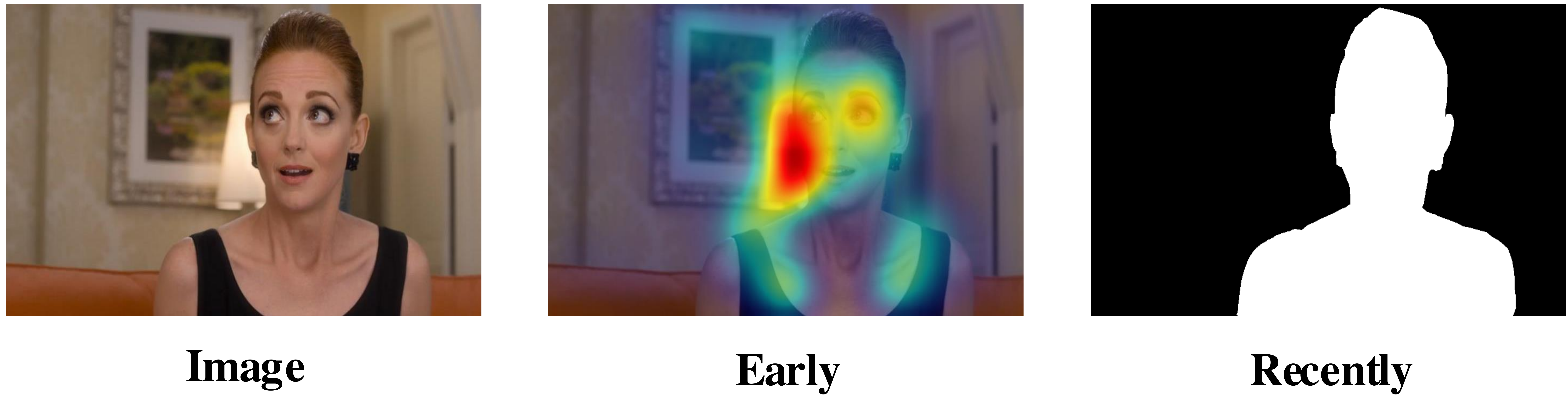}
	\end{center}
	\caption{The introduction of saliency detection.}
	\label{introduce}
\end{figure}

\renewcommand{\shortauthors}{C. Zhu et al.}


\section{Introduction}
Saliency detection is a process of getting a visual attention region precisely from an image. The attention is a behavioral and cognitive process of selectively concentrating on one aspect within an environment while ignoring other things.

An inherent and powerful ability of human eyes is to
quickly capture the most conspicuous regions from a scene, and passes them
to high-level visual cortexes. The attention selection reduces the complexity
of visual analysis and thus makes human visual systems considerably efficient
in complex scenes.

Early works on computing saliency aim to locate visual attention regions. Recently, this field has been extended to locate and refine salient regions and objects as shown in Fig. \ref{introduce}. Serving as a foundation of various multimedia applications, salient object detection has been widely used in content-aware editing~\cite{Chang2011Content,Ji:2012:CSF:2168752.2168758}, image retrieval~\cite{Cheng2014SalientShape,Zhang:2018:RCD:3183892.3158674}, object recognition~\cite{Alexe2012Measuring,leo2017computer}, object segmentation~\cite{Girshick2013Rich,Li2014The}, compression~\cite{Zhu2018An}, image retargeting~\cite{Sun2011Scale,Rao:2016:LHC:2906145.2873065}, and etc.
\begin{figure}[b]
	\begin{center}
		\includegraphics[width=0.8\textwidth]{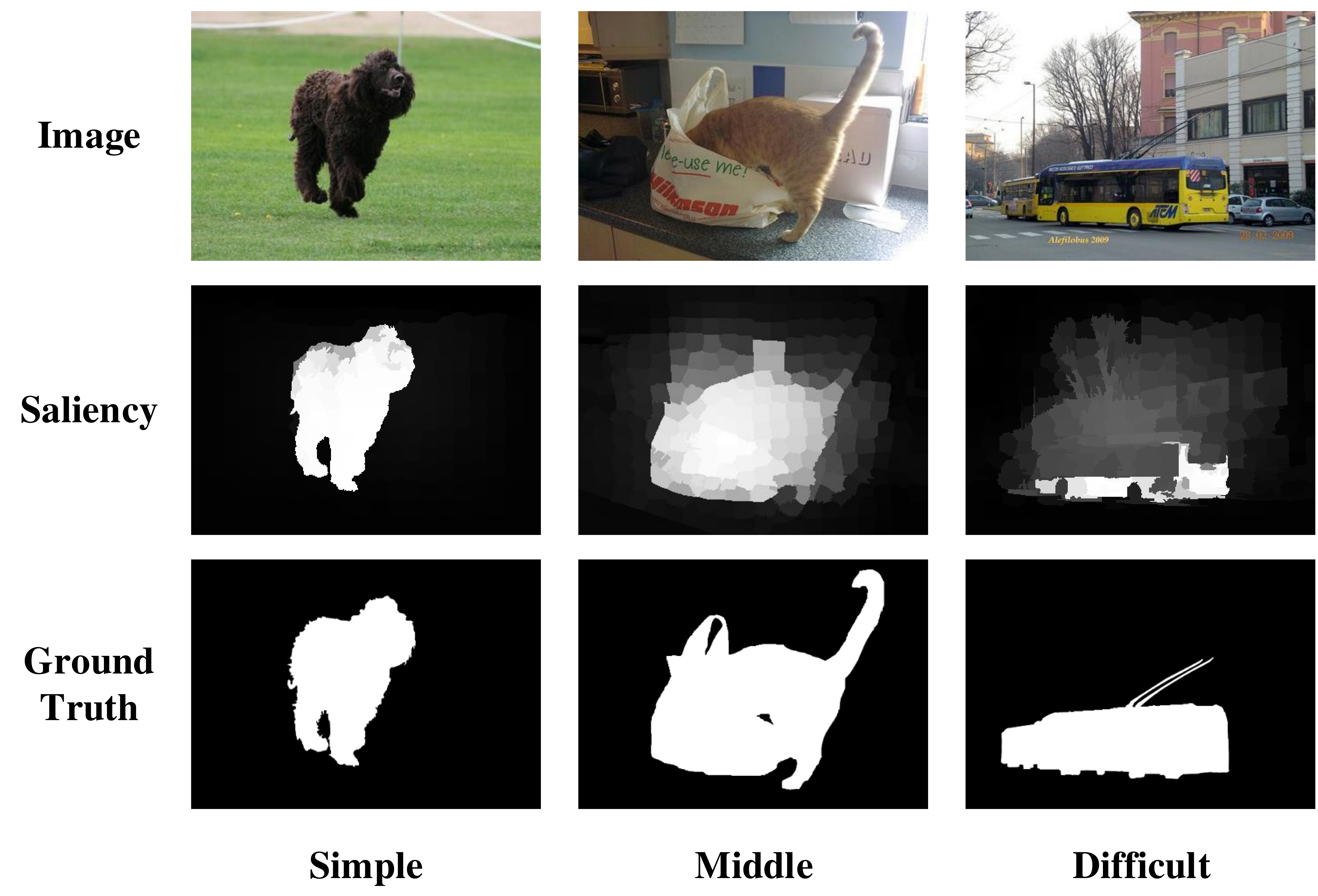}
	\end{center}
	\caption{The illustration of saliency detection with complex scenes. The top row represents input images; the middle row represents saliency detection results with BSCA~\cite{Qin2015Saliency} method; the third row represents the ground truth with the human annotation. From the left to right, the complexity of scenes are gradually increasing. }
	\label{chall}
\end{figure}

In general, saliency detection algorithms mainly use either top-down or bottom-up approaches. Top-down approaches are task-driven and need supervised learning. Bottom-up approaches usually use low-level cues without supervised learning, such as color, distance and other heuristic saliency features. One of the most used heuristic saliency features~\cite{Qin2015Saliency,Achanta2009Frequency,Shi2016Hierarchical,Li2015Inner,Murray2011Saliency,Tong2015Salient,Lu2016Co,Furnari2014An} is the contrast, such as the pixel-based or patch-based, region-based, multi-scale, and center-surround contrasts, and etc. Although those methods have their own advantages, they are not robust in some specific applications and may lead to inaccurate results on challenging salient object detection datasets illustrated in Fig. \ref{chall}.

\begin{figure}
	\begin{center}
		\includegraphics[width=0.8\textwidth]{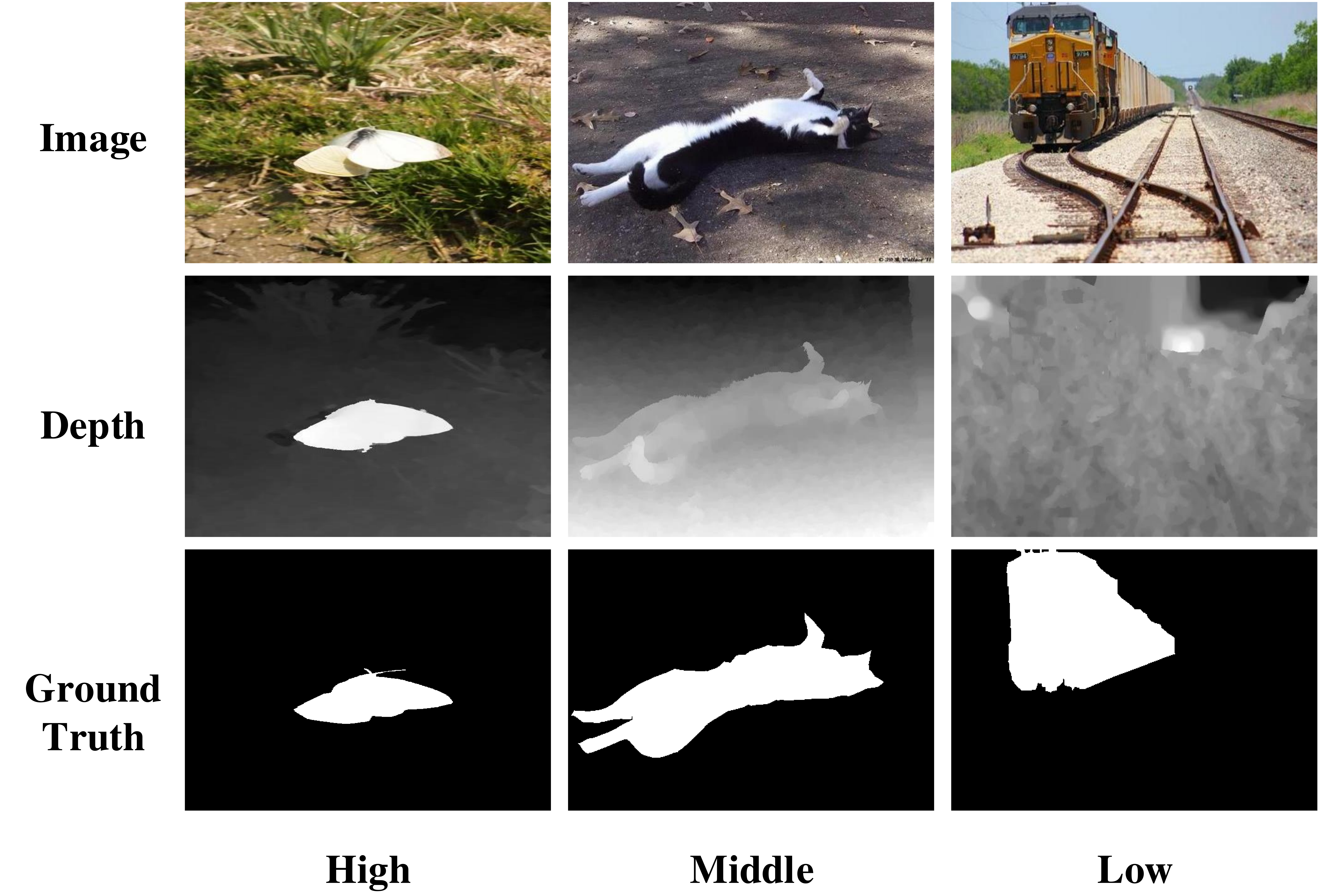}
	\end{center}
	\caption{The illustration of the effects of different contrast depth maps. The top row represents input images; the middle row represents depth maps; the third row represents ground truth with human annotation. And from the left to right, the contrasts of depth maps are gradually decreasing. }
	\label{depthcontrast}
\end{figure}
Recently, advances in 3D data acquisition techniques have motivated the adoption of structural features, improving the discrimination between different objects with similar appearances.
Although, depth cues can enhance salient object regions, it is very difficult to produce good results when a salient object has low depth contrast compared to its background shown in Fig. \ref{depthcontrast}.

\begin{figure}[b]
	\begin{center}
		\includegraphics[width=0.9\textwidth]{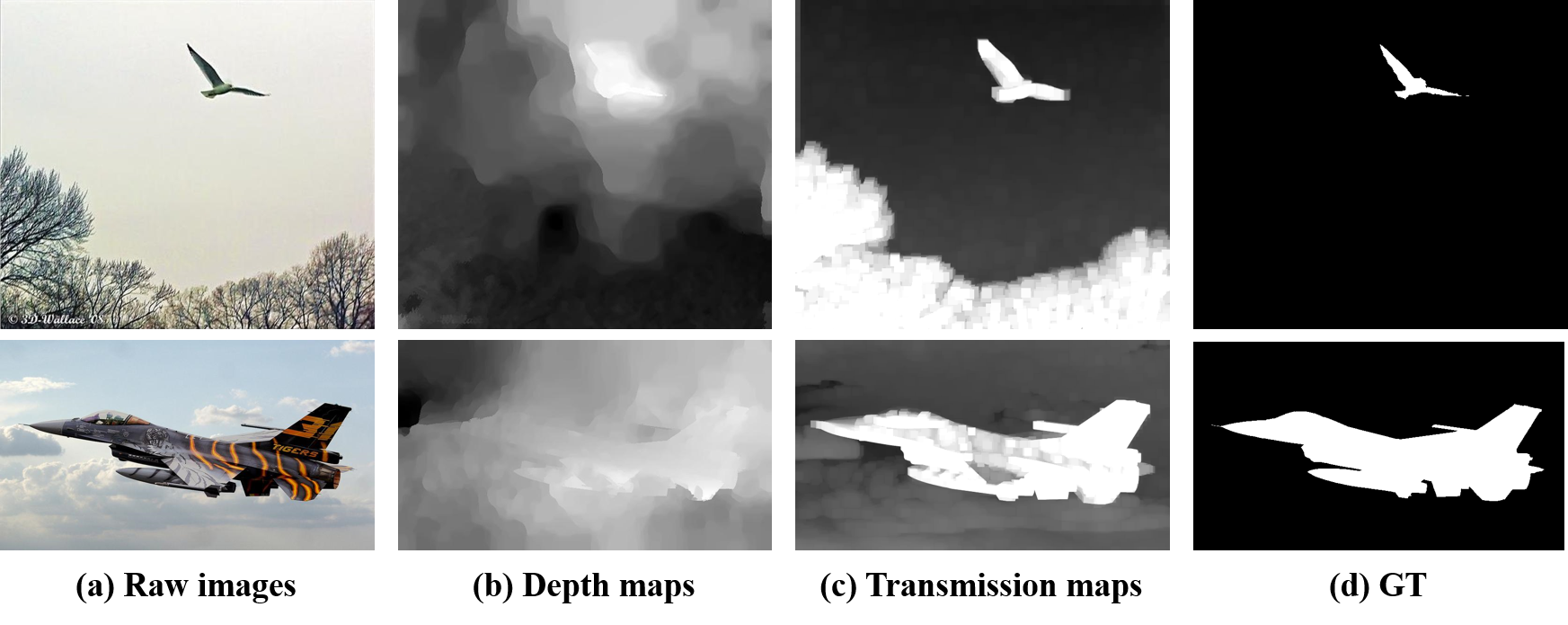}
	\end{center}
	\caption{The illustration of the effect of dark channels, which increases the robustness of detection algorithm when the depth map has the low contrast.}
	\label{trans}
\end{figure}

In order to address the aforementioned difficulties of saliency detection on challenging datasets, we refer to human cognition and use center priors to locate visual saliency and explore the characteristics of salient objects from the perspective of the transmittance. By presenting a new prior knowledge called center-dark channel prior, we can increase the robustness of the detection process and improve the performance of saliency detection algorithm displayed in Fig. \ref{trans}.

\begin{figure}[b]
	\begin{center}
		\includegraphics[width=\textwidth]{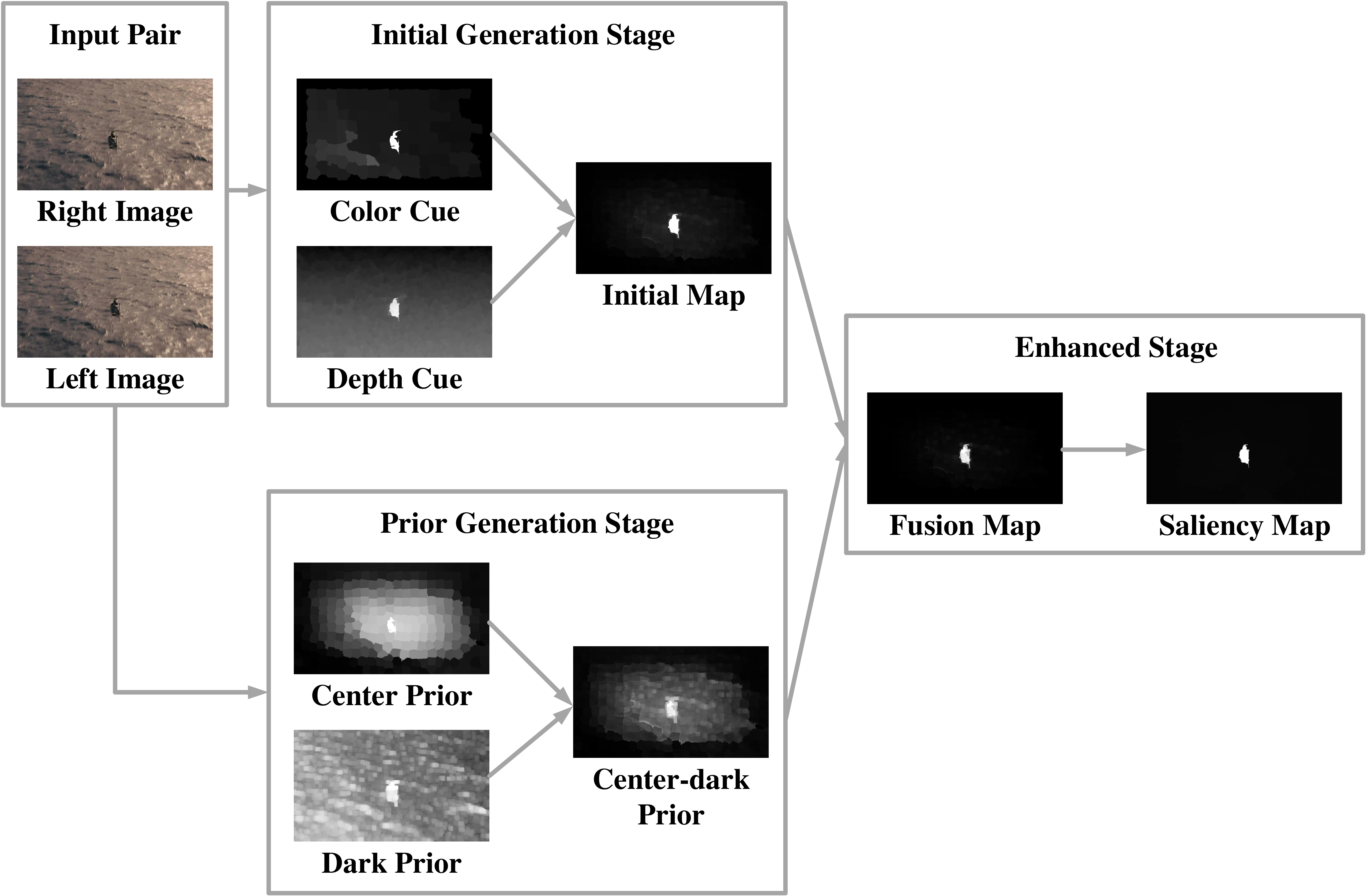}
	\end{center}
	\caption{The framework of the proposed algorithm.}
	\label{framework}
\end{figure}

In this paper, we exploit a new prior, which is called center-dark channel prior, to increase the robustness of saliency detection and propose an innovative saliency detection algorithm using center-dark channel priors demonstrated in Fig. \ref{framework}. First, we generate an initial saliency map based on a color saliency map and a depth saliency map of a given RGB-D image. Second, since salient objects are always located in the center of an image based on cognitive psychology and we also find that the dark channel prior can provide an additional cue for saliency detections, we generate a novel center-dark channel map based on a center saliency prior and a dark channel prior to improve the performance. Finally, we fuse initial saliency maps with center-dark channel priors to generate the final saliency maps.

In summary, major contributions of this work are given as follows:
\begin{itemize}
  \item A novel algorithm is proposed for saliency detection on RGB-D images, where we exploit center-dark channel priors to gain robust detection processes.
  \item Unlike existing works, center-dark channel priors are explored for the first time in the field of saliency detection to enhance performance.
  \item Comparing with previous works, the proposed method achieves dramatical performance improvements on four benchmark datasets.
  \item The proposed method is further studied in the applications of small target detection and demonstrates great performance improvements.
\end{itemize}

To our consideration, this research may help to
figure out visually perceptual characteristics of human visual systems
for images with complex scenes.

\section{Related Work}

In this section, we have a brief review of classical eye fixation models, traditional saliency algorithms, deep learning based saliency detection methods, center prior based saliency detection approaches and dark channel prior based saliency detection methods.
\subsection{Eye Fixation Model}
The first model for saliency prediction is biologically
inspired and based on a bottom-up computational
model that extracts low-level visual features such as intensities,
colors, orientations, textures and motions at multiple
scales. Itti et al.~\cite{Itti1998A} propose a model that combines multiscale
low-level features to create a saliency map. Harel et
al.~\cite{Cerf2008} presents a graph-based alternative that starts from
low-level feature maps and creates Markov chains over various
image maps, treating the equilibrium distribution over
map locations as activation and saliency values.

These models achieve reasonable results. However, the above mentioned models
have limited uses because they frequently do not match actual
human saccades from eye-tracking data. It seems that not only
human attentions are based on low-level features,
but also on high-level semantics ~\cite{Cerf2008} (e.g., faces, humans,
cars, and etc.). Judd et al.~\cite{Judd2009} introduce an approach that
combined low, mid and high-level image features to calculate salient
locations. These features are used in combination with
a linear support vector machine to train a saliency model.
Borji~\cite{Borji2012} also combines low-level features with top-down
cognitive visual features to learn a direct mapping to
eye fixations using Regression, SVM and Ada-Boost classifiers.
\subsection{Traditional Saliency Algorithm}
Saliency detection for conventional images could be implemented
based on either top-down or bottom-up models.
Top-down models~\cite{Oliva2003,Gao2005,Gao2009,Yang2012} require high level interpretation usually provided by training sets in supervised learning. The contextual saliency is formulated
according to the study of visual cognition, where the global scene context of
an image is highly associated with a salient object~\cite{Oliva2003}. The most
distinct features are selected by information theory based methods~\cite{Gao2009}. Salient objects are detected by the joint learning of a dictionary for object features and conditional random field classifiers
for object categorization~\cite{Yang2012}.
While these supervised approaches can effectively detect
salient regions and perform overall better than bottom-up
approaches, it is still expensive to perform the training
process, especially due to time comsuming data collections.

In contrary, bottom-up models~\cite{Qin2015Saliency,Shi2016,ZhuICCV2017,Li2015,7966712} do not require
any prior knowledge, such as object categories, to obtain saliency
maps by using low level features based on center-surround
contrasts. They compute the feature distinctness of a target region,
e.g., pixels, patches or superpixels and then compare to its surrounding regions
locally or globally. For example, the feature difference is computed
across multiple scales, where a fine scale feature map represents
the feature of each pixel while a coarse scale feature map describes
the features of surrounding regions~\cite{Itti1998A}. Also, to compute center-surround
feature contrasts, spatially neighboring pixels are assigned
different weights~\cite{Perazzi2012} or random walks on a graph used in~\cite{Kim2014}.

In addition, most bottom-up models are based on center or boundary priors. The center prior
assumes that foreground salient objects are usually positioned
near the image center and thus assigned high saliency values~\cite{Cheng2013,Luo2011,Margolin2013,Yang2013}. A distance of pixels from the image center
is combined with other features to reduce the contribution
of pixels far from the image center to compute object saliency
~\cite{Cheng2013}. To emphasize the region near the image center, an initial
saliency map is multiplied by a Gaussian distribution centered in an image~\cite{Luo2011}. Multiple Gaussian distribution maps are also employed to
weight the features of pixels adaptively according to the locations
of salient objects in an image~\cite{Margolin2013}. Convex-hull is used to estimate
the center of a salient object when it is not strictly positioned at the
image center~\cite{Yang2013}. However, this assumption puts a strict constraint
on the location of foreground objects in an image, and thus
might not be applicable to various images.
Bottom-up based approaches do not need
data collections and training processes, consequently requiring
little prior knowledge. These advantages make bottom-up
approaches more efficient and easy to implement in a wide
range of real computer vision applications.
A complete survey of these methods is beyond the scope of this paper and we refer readers to a recent survey paper~\cite{Borji2014Salient} for details.
In this paper, we focus on bottom-up approaches.

\subsection{Deep Learning based Saliency Detection Method}
As the performance of deep convolutional neural networks achieving near human-level performances in image classification and recognition tasks, many algorithms adopt deep learning based methods~\cite{zhang2017amulet,wang2017learning,Li2016Deep,He2016Delving,Wang2016Saliency,Li2015Visual,PDNet2018}. Instead of constructing hand-craft features, these kinds of top-down methods have achieved the state-of-the-art performance on many saliency detection datasets. However, deep learning based algorithms have the following limitations: \textit{1) needing a large number of annotated data for training. (2) requiring very time-consuming training in learning processes even with GPU of high computation ability. (3) Obtaining non-uniformly sampled training instances. (4) Suffering sensitivity to noise in training.}

Bottom-up based approaches do not need
data collections and training processes and consequently require
little prior knowledges. These make bottom-up approaches more suitable for real-time applications. Meanwhile, bottom-up methods are not sensitive to image scales, capacities and types. These advantages make bottom-up
approaches even more efficient and easy to saliency detection. In this paper, we
focus on the bottom-up approaches.

\subsection{Center Prior based Methods}
RGB-D saliency computation is a rapidly growing field and offers object detection and attention prediction in a manner that is robust to appearance. Therefore, some algorithms~\cite{Zhu2017Salient,Zhu2017A,Peng2014RGBD,Cheng2014Depth,Geng2012Leveraging} adopt depth cues to deal with the challenging scenarios. In~\cite{Zhu2017Salient}, Zhu et al. propose a framework based on the cognitive neuroscience and use depth cues to represent the depthes of real scenarios. In~\cite{Cheng2014Depth}, Cheng et al. compute salient stimuli in both color and depth spaces. In~\cite{Peng2014RGBD}, Peng et al. provide a simple fusion framework that combines existing RGB-based saliency with new depth-based saliency. In~\cite{Geng2012Leveraging}, Geng et al. define saliency using depth cues computed from stereo images. Their results show that stereo saliency is a useful consideration compared to previous visual saliency analysis. All of them demonstrate the effectivity of depth cues to improve salient object detection.

However, depth cues cannot warrant the robustness of saliency detection when a salient object has a low depth contrast compared to the background. Inspired by cognitive neuroscience, we find that an image always possesses salient objects in its center position. As a result, many algorithms adopt center priors to enhance salient regions. In~\cite{Zhu2017Salient}, Zhu et al. give a framework based on cognitive neuroscience and use center priors to imitate human central fovea. Furthermore, in~\cite{Judd2009Learning}, the contrast against image boundaries is used as a new regional feature to enhance the center position. In~\cite{Qin2015Saliency}, Qin et al. compare boundary clusters with center clusters and then generate different color distinction maps with complementary advantages and integrate them by taking the spatial distance into consideration. All of them demonstrate that the center prior can strengthen saliency regions.

\subsection{Dark Channel Prior based Methods}
The dark channel prior, which was first put forward in~\cite{He2009Single}, is used for single image haze removal. The dark channel prior is based on the statistics of outdoor haze-free images and it is found that the dark pixels often have very low intensities in at least one color channel.
After that, many dehazing methods \cite{kim2013optimized,Berman2016Non,Xie2010Improved} based on dark channel priors are proposed and good results are obtained. Using the dark channel prior is a classic method on calculating the transmission map.
We use the dark channel prior, which can distinguish the foreground from background by the transmission map, to make a more accurately saliency detection.

Partially inspired by the well-known dark-object subtraction technique~\cite{Jr1988An} widely used in multispectral remote sensing systems, we figure out that salient objects tend to have different transmissivities in most regions of the background. As a result, we propose an algorithm to combine dark channel priors with saliency detection results to enhance the performance of saliency detection.

\begin{figure}
	\begin{center}
		\includegraphics[width=\textwidth]{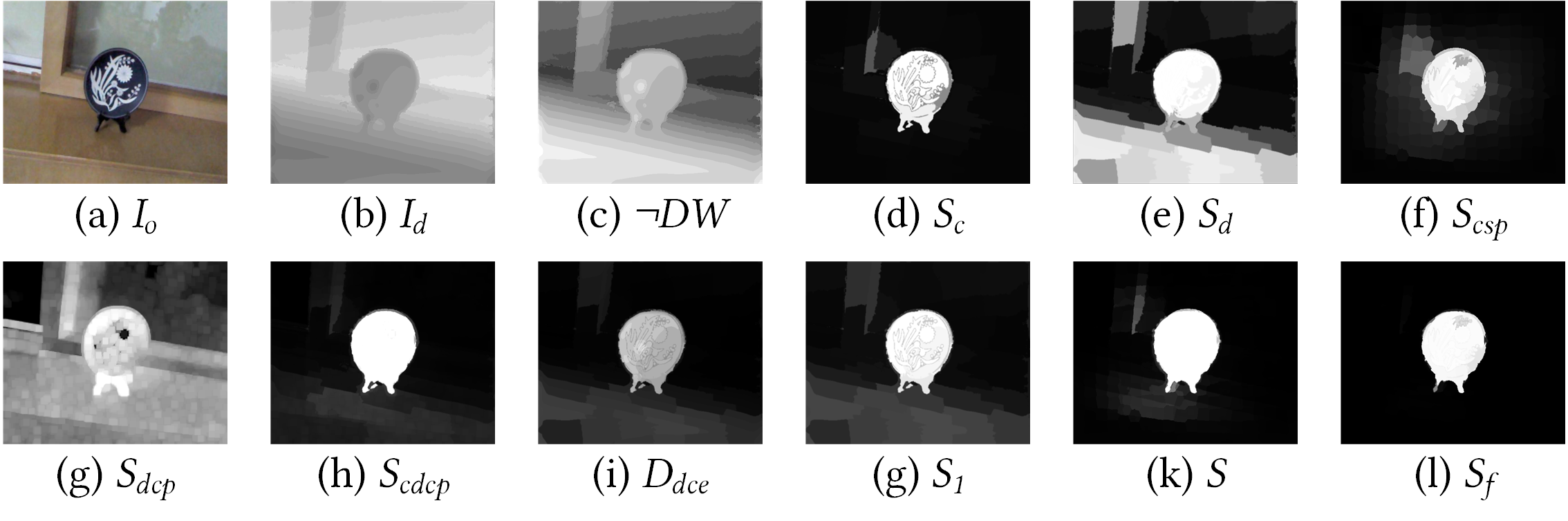}
	\end{center}
	\caption{The visual process of the proposed algorithm.}
	\label{fig:visual process}
\end{figure}

\section{The Proposed Algorithm}
\subsection{Saliency map initialization}
We initialize the saliency map by extracting color and depth features from original image $I_{o}$ and depth map $I_{d}$, respectively.
First, image $I_{o}$ is segmented into $K$ regions based on colors via $K$-means algorithm, which is defined as follows:
\begin{equation}
S_{c}(r_{k}) =\sum_{i=1,i\neq k}^K P_i W_d(r_k) D_c(r_k,r_i),
\end{equation}
where $S_c(r_k )$ is the color saliency of region $k$, $k\in[1,K]$. $r_k$ and $r_i$ represent regions $k$ and $i$, respectively. $D_c (r_k,r_i )$ is Euclidean distance between region $k$ and $i$ in L*a*b color space. $P_i$ represents the area ratio of region $r_i$ compared with whole image. $W_d (r_k )$ is the spatial weighted term of region $k$ as follows:
\begin{equation}
W_d (r_k )=e^{\frac{-D_o (r_k,r_i)}{\sigma^2}},
\end{equation}
where $D_o (r_k,r_i)$ is Euclidean distance between the centers of region $k$ and $i$. $\sigma$ is the parameter controlling the strength of $W_d (r_k)$.

Similar to the color saliency, we define:
\begin{equation}
S_d (r_k )=\sum_{i=1,i\neq k}^K P_i W_d (r_k ) D_d (r_k,r_i ),
\end{equation}
where $S_d (r_k)$ is depth saliency of $I_d$. $D_d (r_k,r_i)$ is Euclidean distance between region $k$ and $i$ in depth space.

In most cases, a salient object is always located in the center of an image or close to a camera center. Therefore, we assign the weights to both center-bias and depth for both colors and depth images. The weights of region k are as assigned by:
\begin{equation}
W_{cd} (r_k)=\frac{ G(\parallel P_k-P_o \parallel )}{N_k}DW(d_k ),
\end{equation}
where $G(\cdot)$ represents Gaussian normalization. $\|\cdot\|$ is Euclidean distance. $P_k$ is the position of region $k$. $P_o$  is the center position of this map. $N_k$ is the number of pixels in region $k$. $DW(d_k)$ is the depth weight, which is given as follows:
\begin{equation}
DW(d_k )=(max\{d\}-d_k )^\mu,
\end{equation}
where $max\{d\}$ represents the maximum depth of the image. $d_k$ is the depth value of region $k$. $\mu$ is a fixed value for a depth map formulated by:
\begin{equation}
\mu=\frac{1}{max\{d\}-min\{d\}},
\end{equation}
where $min\{d\}$  represents the minimum depth of the image.

Then, the initial saliency value of region $k$ is calculated by the following equation:
\begin{equation}
S_1 (r_k )=G(S_c (r_k )W_{cd} (r_k )+S_d (r_k )W_{cd} (r_k )).
\end{equation}

\subsection{Center-dark channel prior}
By mixing the center saliency and the dark channel priors to generate a new prior, we define this prior as the center dark channel prior. By using this prior, we can get a more accurate saliency map.
\begin{figure}[t]
	\begin{center}
		\includegraphics[width=\textwidth]{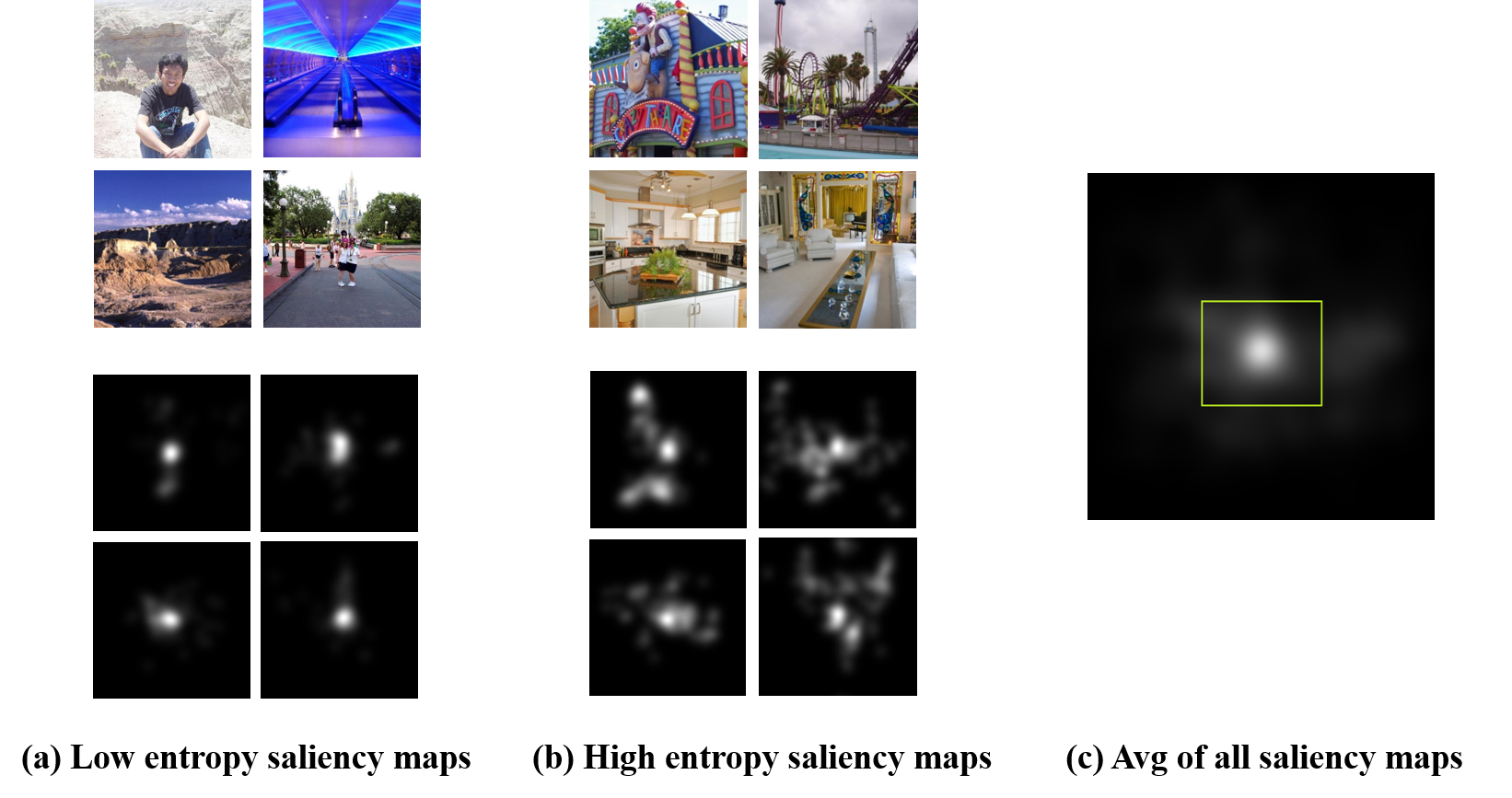}
	\end{center}
	\caption{ Proof of Center Prior: (a) and (b) show examples of saliency maps made from human fixations with low and high entropy and their corresponding images. (c) is a plot of all the saliency maps from human eye fixations indicating a strong bias to the center of the image. 70\% of fixations lie within the indicated rectangles. }
	\label{fig:validation on center prior }
\end{figure}


{\bf Center saliency prior.} According to cognitive neuroscience, human eyes use central fovea to locate objects and make them clearly visible. Therefore, most of the images taken by cameras always locate salient objects around the center. It was shown in \cite{Judd2009Learning} that a saliency map based on the distance of each pixel to the center of the image provides a better prediction of salient objects than many previous saliency models, as shown in Fig. \ref{fig:validation on center prior }.
Inspired by this observation, we further incorporate a center prior into our saliency estimation.

\begin{figure}[b]
	\begin{center}
		\includegraphics[width=0.9\textwidth]{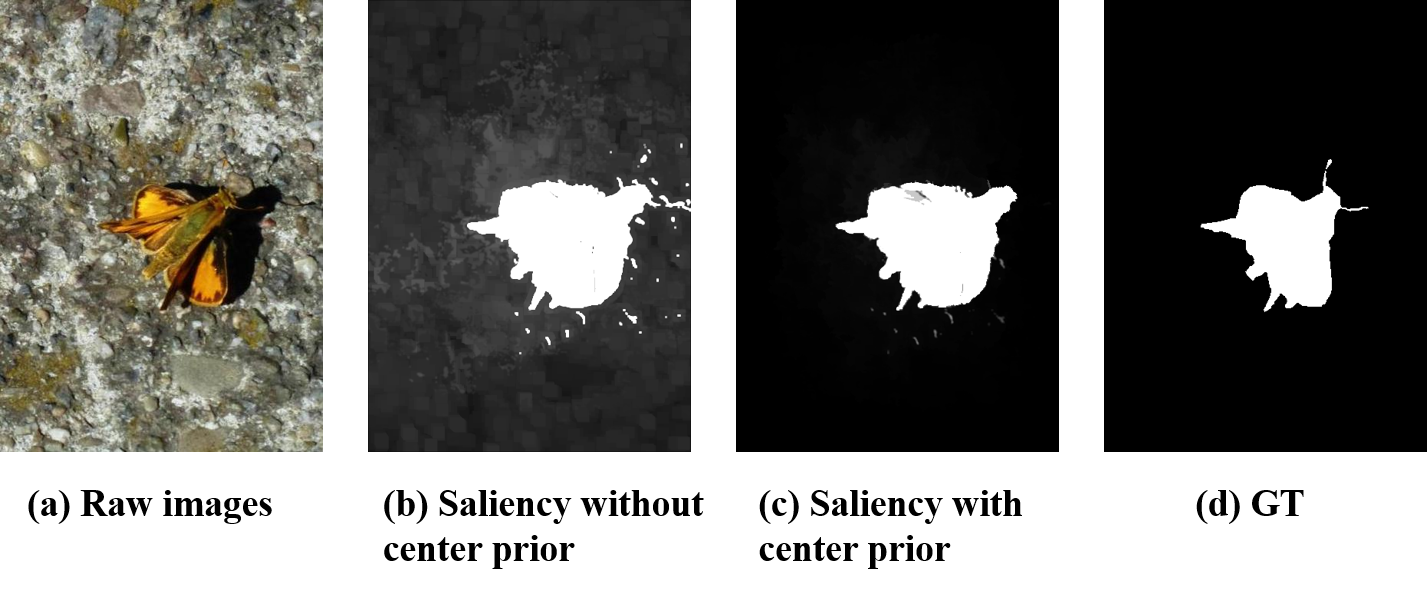}
	\end{center}
	\caption{ Comparing saliency results without (b) and with (c) the center prior, and (d) is the ground truth.}
	\label{fig:center prior}
\end{figure}

Fig. \ref{fig:center prior} provides a visual comparison between saliency
maps obtained with and without center priors and more accurate results
are obtained when the center prior is included.

To get the center saliency map, we use the BSCA algorithm~\cite{Qin2015Saliency}. It constructs the global color distinction and the spatial distance matrix based on clustered boundary seeds and integrates them into a background-based map. Thus, it can improve the accuracy of center objects by erasing the image edges' effects. As shown in the Fig. \ref{fig:visual process}(f), the center saliency map can remove the surroundings area and reserve most of salient regions of an image. We denote this center-bias saliency map as $S_{csp}$.

{\bf Dark channel prior.} The dark channel prior is a popular prior which is widely used in the image haze removal field~\cite{He2009Single}. It is based on the statistics of outdoor haze-free images. The dark channel can detect the most haze-opaque region and improve the atmospheric light estimation. Inspired by dark channel priors, we find that the foreground and background have different transmissivities, so we can distinguish salient objects from backgrounds as shown in Fig. \ref{fig:visual process}(g). We apply this theory in the field of saliency detection and denote the transmissivity map of the dark channel prior as $S_{dcp}$.

\begin{figure}[t]
	\begin{center}
		\includegraphics[width=\textwidth]{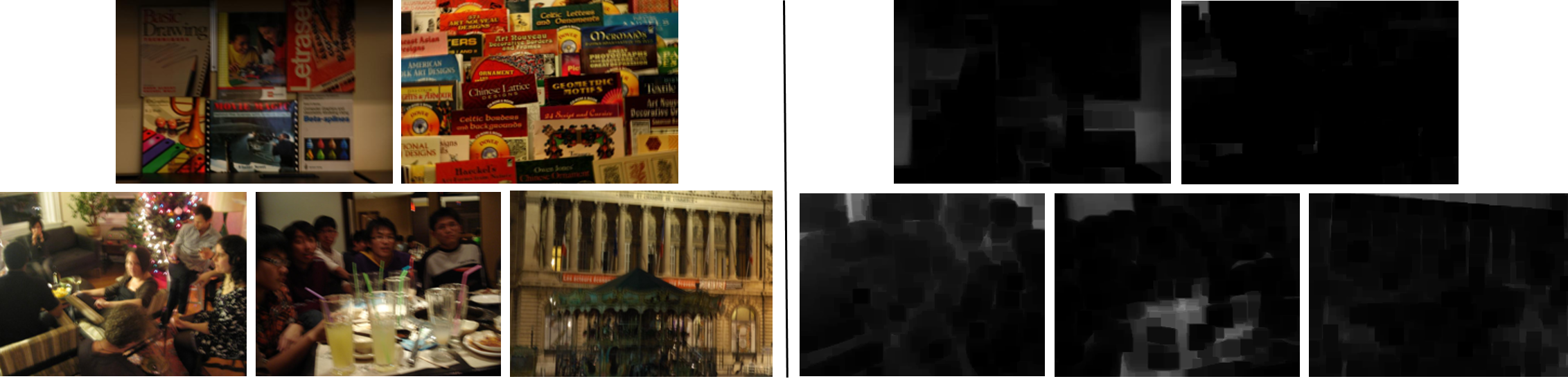}
	\end{center}
	\caption{From left to right: example images and corresponding dark channels.}
	\label{fig:dark channel image}
\end{figure}

Based on the statistics of outdoor haze-free images, He et al. \cite{He2009Single} find that, for the patches of general images, at least one color channel has very low intensity. In other words, the minimum intensity of a general image should has a very low value.Formally, for an image $J$, we define
\begin{equation}
J^{dark}(x) = \min_{c\in{r,g,b}}(\min_{y\in\Omega(x)}(J^c(y)) ) \approx 0
\end{equation}
where $J_{c}$ is a color channel of $J$ and $\Omega(x)$ is a local patch centered at $x$. We call $J^{dark}$ the dark channel of $J$.  According to the experiments of the dark channel processing on general images, we find that the dark channel image tends to be zero, as provided in Fig. \ref{fig:dark channel image}.

\begin{equation}
\hat{t} (x)=1-\underset{c\in{r,g,b}}{min}\left \{ \underset{y\in {\Omega(x)}}{min}\left\{\frac{I^c(y)}{A^c}\right \} \right \}.
\end{equation}

\begin{figure}[b]
	\begin{center}
		\includegraphics[width=0.9\textwidth]{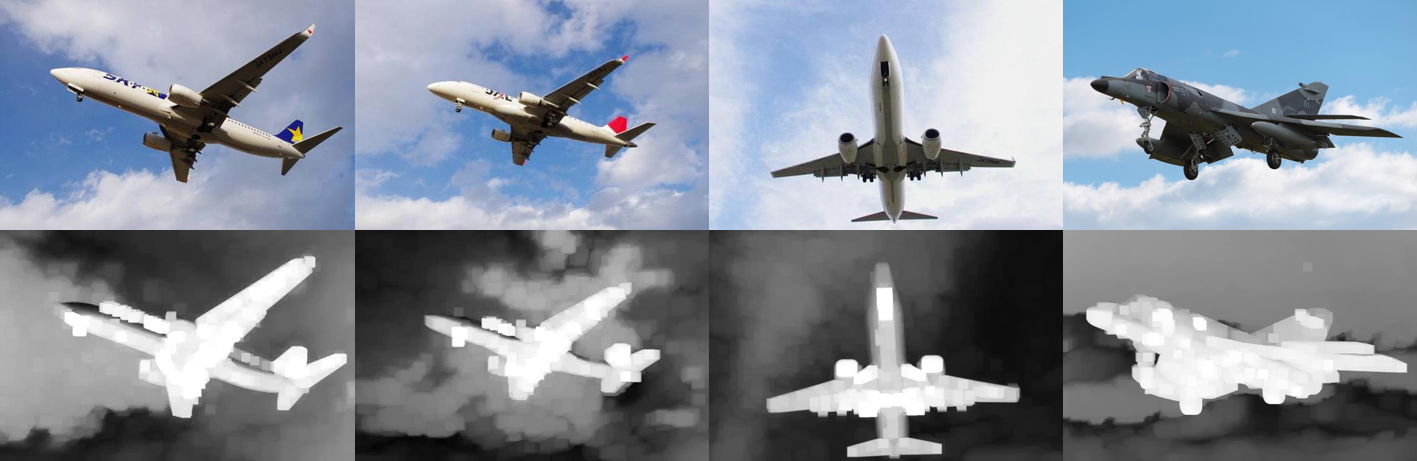}
	\end{center}
	\caption{ Top row: example images; Bottom row: the corresponding transmission maps. }
	\label{fig:dcp}
\end{figure}

According to \cite{He2009Single}, the estimated transmission value can be calculated using Eq. (9), where $I_{c}$ is a color channel of $I$ and $I$ is an image with haze. $A_{c}$ is the global light which is constant for most images.
The transmissivity of an image is the reverse value of dark channel results. Since the objects that are closer to the camera have less photographic fog, the foreground part will be the haze\_free part and the dark channel result tends to be zero. So the transmissivity of the foreground is larger than that of the background as illustrated in Fig. \ref{fig:dcp}.

\subsection{Saliency refinement with an updated fusion}
Based on the initialized saliency map, we utilize an updated fusion to refine saliency maps. The updated fusion consists of the depth cue, the center-dark channel prior and the updated fusion based enhancements.

{\bf Depth cue based enhancement.} The depth cue makes salient objects prominent. We denote it by:
\begin{equation}
D_{dce} (d_k )=Norm(\neg DW(d_k )),
\end{equation}
where $\neg$ is the negation operation which can enhance the saliency degree of front regions as shown in Fig. 2(c), because the foreground object has a low depth value while the background one possesses a high depth value in depth map. $Norm(\cdot)$ is a normalized operation.

{\bf Center-dark channel prior based enhancement.} We combine the center saliency and the dark channel priors to enhance the final saliency results denoted as follows:
\begin{equation}
S_{cdcp} (r_k )=Norm(S_{csp} (r_k )) Norm(S_{dcp} (r_k )),
\end{equation}

{\bf Updated fusion based enhancement.} We fuse the depth cue and the center-dark channel prior based enhancements with the initial saliency values as shown below:
\begin{equation}
S(r_k )=(1-e^{-(S_1 (r_k ) + D_{dce} (d_k ) + S_{cdcp} (r_k ))}) S_1 (r_k )S_{csp} (r_k ),
\end{equation}
where $S(r_k )$ is the fusion enhanced saliency value and $S_1 (r_k )$ is the initial saliency value denoted in Section 2.1.

To refine saliency results, we updated the fusion enhanced saliency value in the following equation:
\begin{equation}
S_f(r_k )=1-e^{-(S_1 (r_k ) S_{csp} (r_k )S(r_k ))}.
\end{equation}
where $S_f (r_k )$ is the final saliency value.

From the Fig. \ref{fig:visual process}, we can see the visual results of the proposed algorithm. The main steps of the proposed salient object detection algorithm are summarized in Algorithm 1.
\begin{algorithm}[!htb] %
	\caption{ The proposed saliency algorithm using centre-dark channel prior} %
\begin{flushleft}
{\bf Input:} original maps $I_o$, depth maps $I_d$;\\	
{\bf Output:} final saliency values $S_f$;
\end{flushleft}
	\begin{algorithmic}[1] %
		\STATE {\bf for} each region $k=1,K$ {\bf do:} \\
    		\STATE \quad compute color saliency values $S_c (r_k )$ and depth saliency values $S_d (r_k )$;
    		\STATE \quad calculate center-bias and depth weights $W_{cd} (r_k )$;
    		\STATE \quad get initial saliency value $S_1 (r_k )$;
		\STATE {\bf end for}
		\STATE obtain center saliency priors $S_{csp}$ and dark channel priors $S_{dcp}$;
		\STATE figure out final saliency values $S_f$ by updated fusion;
		\RETURN final saliency values $S_f$; %
	\end{algorithmic}
\end{algorithm}

\section{Experimental Evaluation}
\subsection{Datasets}
In this section, we evaluate the proposed method on four RGB-D datasets.

{\bf NJU2000}~\cite{Ju2015Depth}. The NJUDS2000 dataset contains 2000 stereo images as well as the corresponding depth maps and manually labeled ground truths. The depth maps are generated using an optical flow method.

{\bf NLPR}~\cite{Peng2014RGBD}. The NLPR RGB-D salient object detection dataset contains 1000 images captured by Microsoft Kinect in
various indoor and outdoor scenarios.


{\bf RGBD135}~\cite{Cheng2014Depth}. This dataset has 135 indoor images taken by Kinect with the resolution $640 \times 480$.

{\bf SSD100}~\cite{8265566}. This dataset is built on three stereo movies. It contains 80 images with both indoors and outdoors scenes.

\subsection{Evaluation Metrics}
Four most widely used evaluation metrics are used to serve as the performance measurements of saliency algorithms, including the precision-recall (PR) curves, F-measure, receiver operating characteristics (ROC) curve and mean absolute error (MAE).

The precision is defined as follows:
\begin{equation}
Precision=\frac{\|{p_i\mid d(p_i)\geq d_t}\cap{p_g}\|}{\|{p_i\mid d(p_i)\geq d_t}\|} ,
\end{equation}
where ${p_i\mid d(p_i)\geq d_t}$ indicates the set that binarized from a saliency map using threshold $d_t$. ${p_g}$ is the set of pixels belonging to ground truth salient objects.

The recall is defined below:
\begin{equation}
Recall = \frac{\|{p_i\mid d(p_i)\geq d_t}\cap{p_g}\|}{\|{p_g}\|} .
\end{equation}
The precision-recall curve is plotted by connecting P-R scores for all thresholds.

The F-measure is a weighted average of precision and average recall and can be calculated by the following formula:
\begin{eqnarray}
{F}_{\beta} = \frac{(1+{\beta}^{2}) \cdot Precision \cdot Recall}{{\beta}^{2}\cdot Precision + Recal}.  \nonumber \\
\end{eqnarray}
Here, as suggested by previous works, we set ${\beta}^{2}$ to 0.3 to emphasize the precision rather than recall.

The ROC curve is the plot of true positive rates (TPR) versus false positive rates (FPR) by testing all possible thresholds.

The MAE captures the average difference between the produced saliency map and the ground truth map and it is expressed as:

\begin{eqnarray}
MAE = \frac{1}{W\times H} \sum_{x=1}^{W} 
\end{eqnarray}
where G is the binary ground truth mask. W and H are width and height of the saliency map S,respectively.

\subsection{Ablation Study}
We first validate the effectiveness of each step in our method: initial saliency detection, depth cue enhanced saliency detection, center-dark channel prior saliency detection, updated fusion saliency detection and final saliency detection. Table. 1 shows the validation results on NLPR and RGBD135 datasets. The accumulated processing gains can clearly be seen after each step and the final saliency results shows the good performance. After all, it proves that each step in our algorithm is effective for generating the final saliency maps.
\begin{table}[!h]
  \centering
\caption{Ablation study results on two RGB-D datasets.}
\label{ablation}
\begin{tabular}{|c|c|c|c|c|}
    \hline
    \multirow{2}{*}{}&
    \multicolumn{2}{c|}{NLPR} &
    \multicolumn{2}{c|}{RGBD135}\\ \cline{2-5}
    &${F}_{\beta}$ &MAE&${F}_{\beta}$&MAE\\
    \hline
    $S_1$ & 0.6497& 0.1978& 0.6528 &0.2004 \\ \hline

    $D_{dce}$ &0.6725 &0.1528 &0.6785 &0.1685 \\ \hline

    $S_{cdcp}$ &  0.6764 &  0.1442 &  0.6889 & 0.1489     \\  \hline

    $S$  &   0.6875  &  0.1206   & 0.6958& 0.1343 \\ \hline

    $S_f$  &   0.7056  & 0.0860  &  0.7105   &  0.0794   \\  \hline

\end{tabular}
\end{table}


\subsection{Comparison}
To further illustrate the effectiveness of our algorithm, we compare our proposed methods with FT~\cite{Achanta2009Frequency}, SIM~\cite{Murray2011Saliency}, HS~\cite{Shi2016Hierarchical}, BSCA~\cite{Qin2015Saliency}, LPS~\cite{Li2015Inner}, DES~\cite{Cheng2014Depth}, NLPR~\cite{Peng2014RGBD}. We use the codes provided by the authors to reproduce their experiments. For all the compared methods, we use the default settings suggested by authors. And for the Eq. 2, we take $\sigma^2 = 0.4$ which has the best contribution to the results.

\begin{figure}
	\begin{center}
		\includegraphics[width=\textwidth]{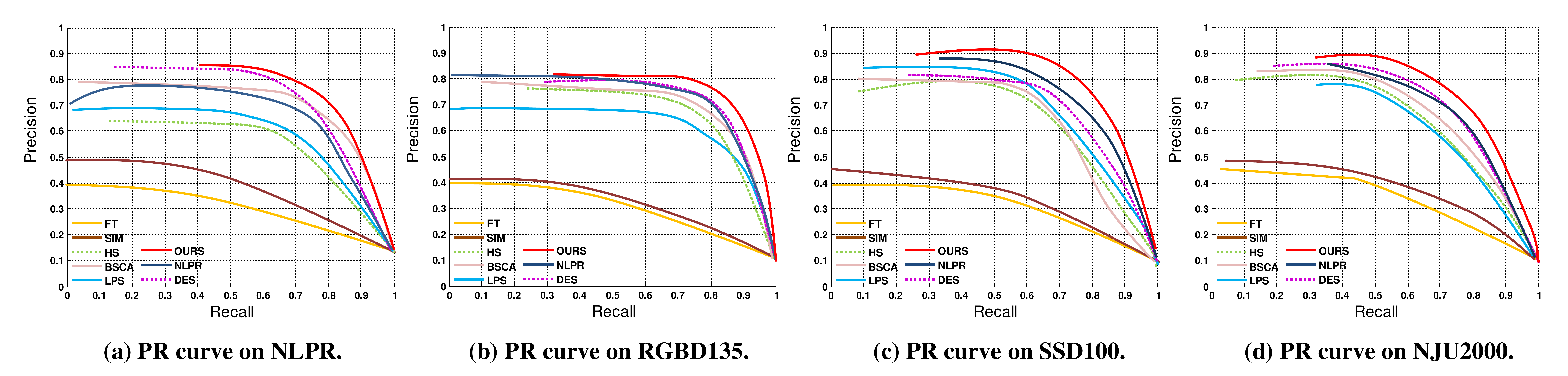}
	\end{center}
	\caption{PR curves of different methods on four datasets. }
	\label{fig:PR curve}
\end{figure}

\begin{figure}
	\begin{center}
		\includegraphics[width=\textwidth]{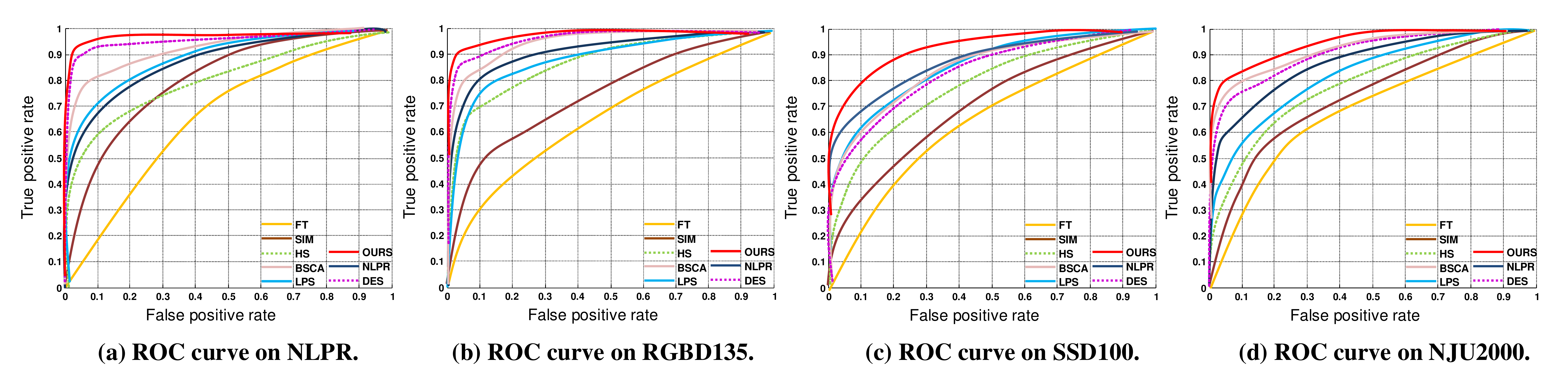}
	\end{center}
	\caption{ROC curves of different methods on four datasets. }
	\label{fig:ROC curve}
\end{figure}

The precision and recall evaluation results and ROC evaluation results are shown in Fig. \ref{fig:PR curve} and Fig. \ref{fig:ROC curve}, respectively.
\begin{figure}
	\begin{center}		
		\includegraphics[width=\textwidth ]{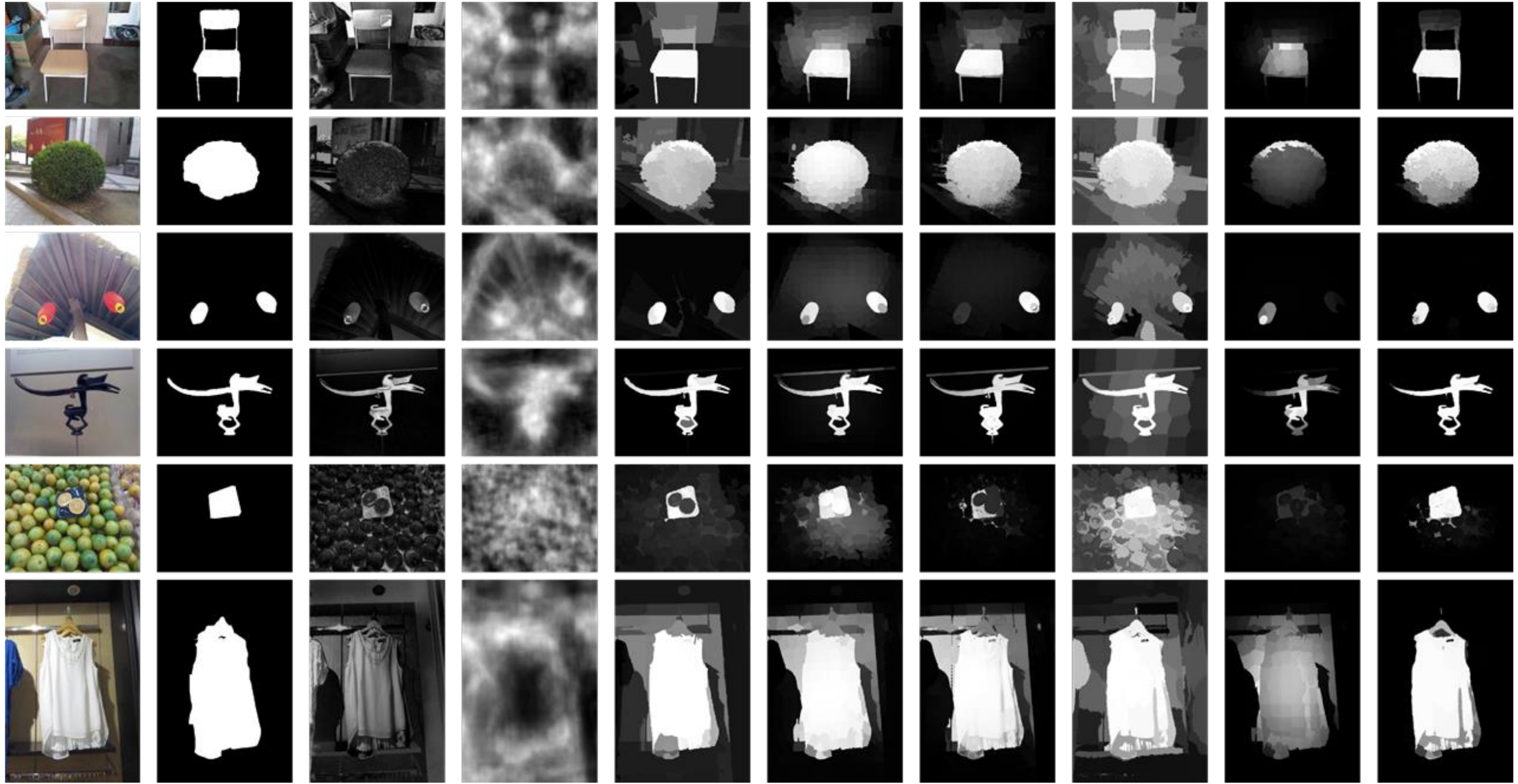}
	\end{center}
	\caption{Visual comparison of saliency maps on four datasets. From left to right: original images, ground truth, FT, SIM, HS, BSCA, LPS, DES, NLPR and OURS, respectively.}
	\label{fig:visual comparison}
\end{figure}

From the precision-recall and ROC curves, we can see that our saliency detection method can achieve better results on all four datasets.

MAE results on four datasets are listed in Table 2, where the lower value generates the better performance. And F-measure results on four datasets are provided in Table 3, where the higher value produces the better performance. The best results are emphasized in boldface. Comparing with the MAE values and F-measure values, it can be observed that our saliency detection method is superior and can obtain the most precise salient regions among tested approaches.
\begin{table}
	\begin{center}
		\begin{tabular}{|c|c|c|c|c|}
			\hline
			 & RGBD135 Dataset & NLPR Dataset & SSD100 Dataset & NJU2000 Dataset \\
			\hline
			FT &0.2049 &0.2168 & 0.2738 &0.2637\\
			\hline
			SIM &0.3740 &0.3957 & 0.4184 &0.4012\\
			\hline
			HS &0.1849 &0.1909 & 0.2582 &0.2516\\
			\hline
			BSCA &0.1851 &0.1754 & 0.2386 &0.2148\\
			\hline
			LPS &0.1406 &0.1252 & 0.1960 &0.2059\\
			\hline
			DES &0.3079 &0.3207 & 0.3132 & 0.4465\\
			\hline
			NLPR &0.1165 &0.1087 & 0.1784 & 0.1669\\
			\hline
			OURS &{\bf 0.0794} &{\bf 0.0860} &{\bf 0.1779} &{\bf 0.1589}\\
			\hline
		\end{tabular}
	\end{center}
	\caption{MAE evaluation results. The best results are shown in boldface. }
\end{table}

\begin{table}
	\begin{center}
		\begin{tabular}{|c|c|c|c|c|}
			\hline
			 & RGBD135 Dataset & NLPR Dataset & SSD100 Dataset & NJU2000 Dataset \\
			\hline
			FT &0.4361 &0.4488 &0.4687 & 0.4723\\
			\hline
			SIM &0.4411 &0.4525 &0.4889 & 0.4912\\
			\hline
			HS &0.5361 &0.6003 &0.5716 & 0.6090\\
			\hline
			BSCA &0.5826 &0.5925 &0.5755 &0.6290\\
			\hline
			LPS &0.5452 &0.5890 &0.5935 &0.5692\\
			\hline
			DES &0.5410 &0.5915 &0.5797 &0.6202\\
			\hline
			NLPR &0.4912 &0.5957 &0.6415 &0.6165\\
			\hline
			OURS &{\bf 0.7105} &{\bf 0.7056} &{\bf 0.7212}&{\bf 0.7198}\\
			\hline
		\end{tabular}
	\end{center}
	\caption{F-measure evaluation results. The best results are shown in boldface. }
\end{table}

The visual comparisons are given in Fig. 4 and the advantages of our method are clearly demonstrated. Our method can detect both a single salient object as well as multiple salient objects more precisely. In contrast, the compared methods may fail in some situations.

\section{Small Target Detection}
It is very interesting to find that the proposed saliency detection algorithm using center-dark channel priors is also valid in small target detection.
\begin{figure}
	\begin{center}
		\includegraphics[width=\textwidth,]{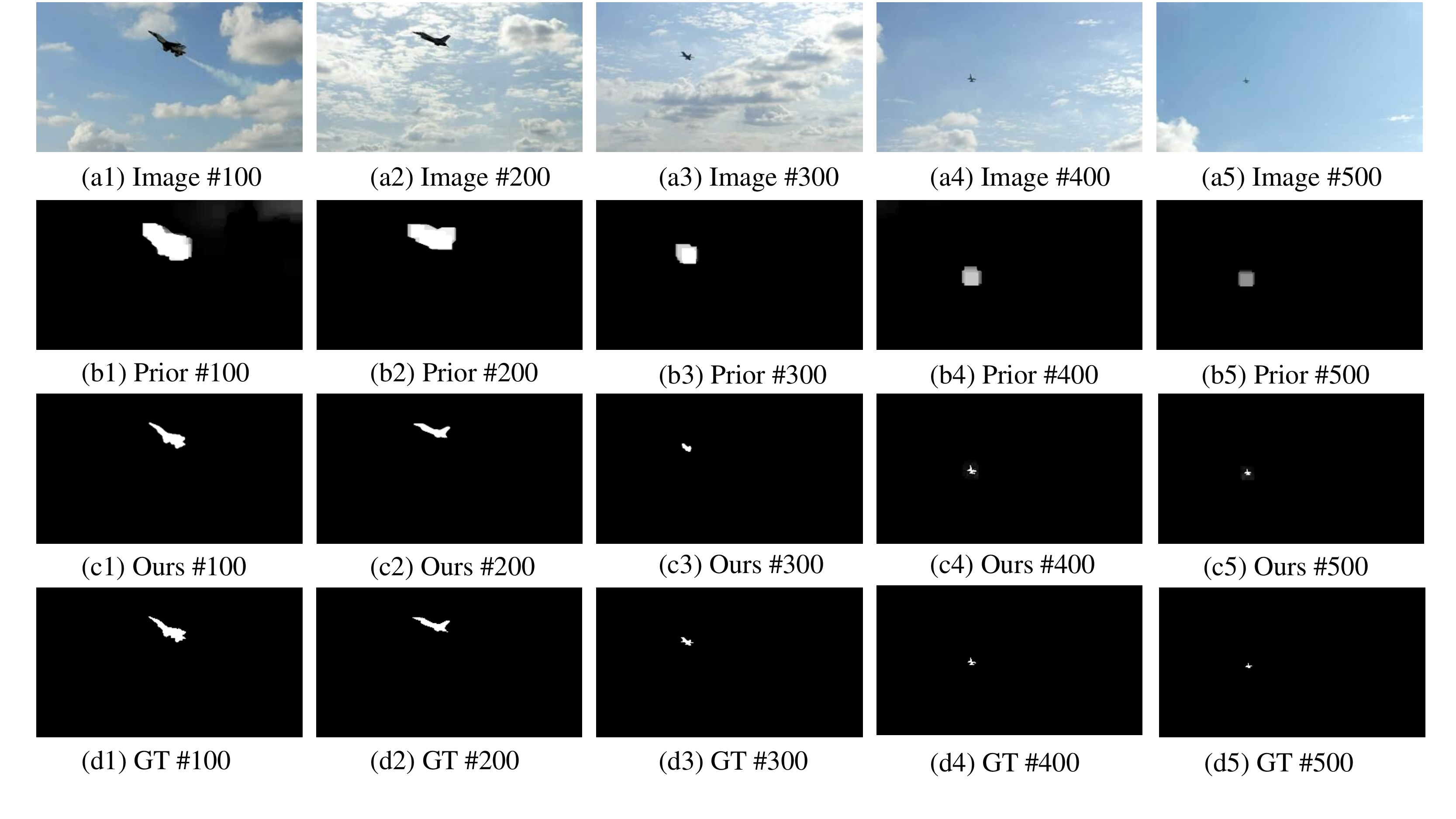}
	\end{center}
	\caption{The proposed algorithm is applied in small target detection. (a1)-(a5) represent different frames of original videos.(b1)-(b5) represent different frames of the proposed priors detection results.(c1)-(c5) represent different frames of the proposed algorithm detection results. (d1)-(d5) represent different frames of the ground truth. }
	\label{small}
\end{figure}

Small target detection plays an important role in many computer vision tasks, including early warning systems, remote sensing and visual tracking.
Different from traditional objects detection, small target detection is usually more difficult since the resolution is low and the feature is fuzzy.
Meanwhile, several factors such as sensor noises, size variations and artificial interferences make the task even more challenging.
Over the past decades, numerous small targets detection models have been proposed. Their strategies can be divided into three main categories: target enhancement
\cite{zhu2015effective}, background suppression \cite{bae2012edge}, and figure-ground segregation \cite{chen2014local}. With the advantage of simultaneously enhancing target signals and suppressing background clutters, the third
category of methods usually exploits different contrast techniques to model this task. In
computer vision, the contrast mechanisms, including local center-surround difference and
global rarity, are closely related to human visual perception and widely used in bottom-up
saliency detection models. Since the visual saliency which stemmed from psychological science
has attracted more attention, some saliency map based methods
are also proposed in recent years \cite{lee2015effective,qi2013robust,li2009saliency}.

In many cases, small targets are immersed in cloud clutters. For outdoor scenes, dark channel priors can be used to distinguish sky background easily and capture objects flying in sky. Combining with saliency detection, we can gain a better result.
We present parts of our experimental results on these challenging dataset~\cite{Lou2016Small}.
The experimental results by applying the proposed algorithm to small target detections are shown in Fig. \ref{small} to support our claims. The reason behind why the proposed algorithm can be transplanted in small target detections is that (1) the proposed center-dark channel prior can be used to locate small objects and (2) the proposed saliency detection algorithm can be applied to refine small targets features. Therefore, we claim that the proposed center-dark channel prior can be used for small target detections in addition to saliency detections.

\section{Discussion and Future Directions}
\subsection{Fail Case}
\begin{figure}
	\begin{center}
		\includegraphics[width=\textwidth]{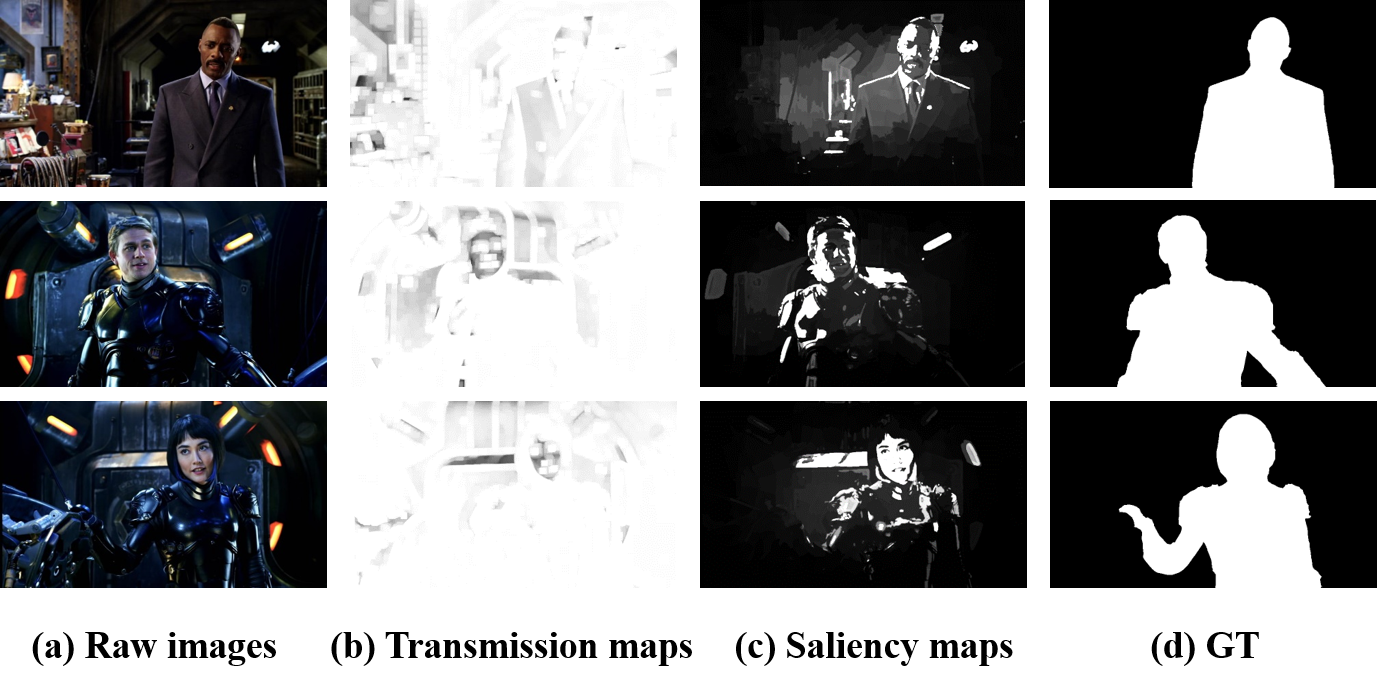}
	\end{center}
	\caption{The fail cases for imaging in dark environment.}
	\label{fail case}
\end{figure}

For dark channel priors, it is more suitable for outdoor scenes to emphasize the brightness of the background.
So when the background is too dark, the transmission map of images will be difficult to distinguish foregrounds from backgrounds as shown in Fig \ref{fail case}(b). The resulting saliency map is shown in Fig \ref{fail case}(c) and we can see that the borders are not clear.

\subsection{Future Directions}
In the past two decades, hundreds of algorithms have been proposed for saliency detection.
On all these commonly-used datasets, top-down algorithms can achieve wonderful precision (over 80$\%$). However, the performance improvement has encountered a bottleneck. In order to solve this challenging issue, we suggest the following ideas based upon our research experience:

(1) More innovative researches should be conducted based upon cognitive science, such as cognitive mechanism based algorithms.

(2) More multi-level annotated datasets should be built. Since almost all datasets only provide rough binary masks as ground truth, we suggest that ground truth maps should be annotated in multiple levels. Different labeled levels correspond to different levels of the saliency.

(3) More diversified medias should be introduced into this field, such as speech and text based saliency detection.

\begin{figure*}
  \centering
 \includegraphics[width=\textwidth ]{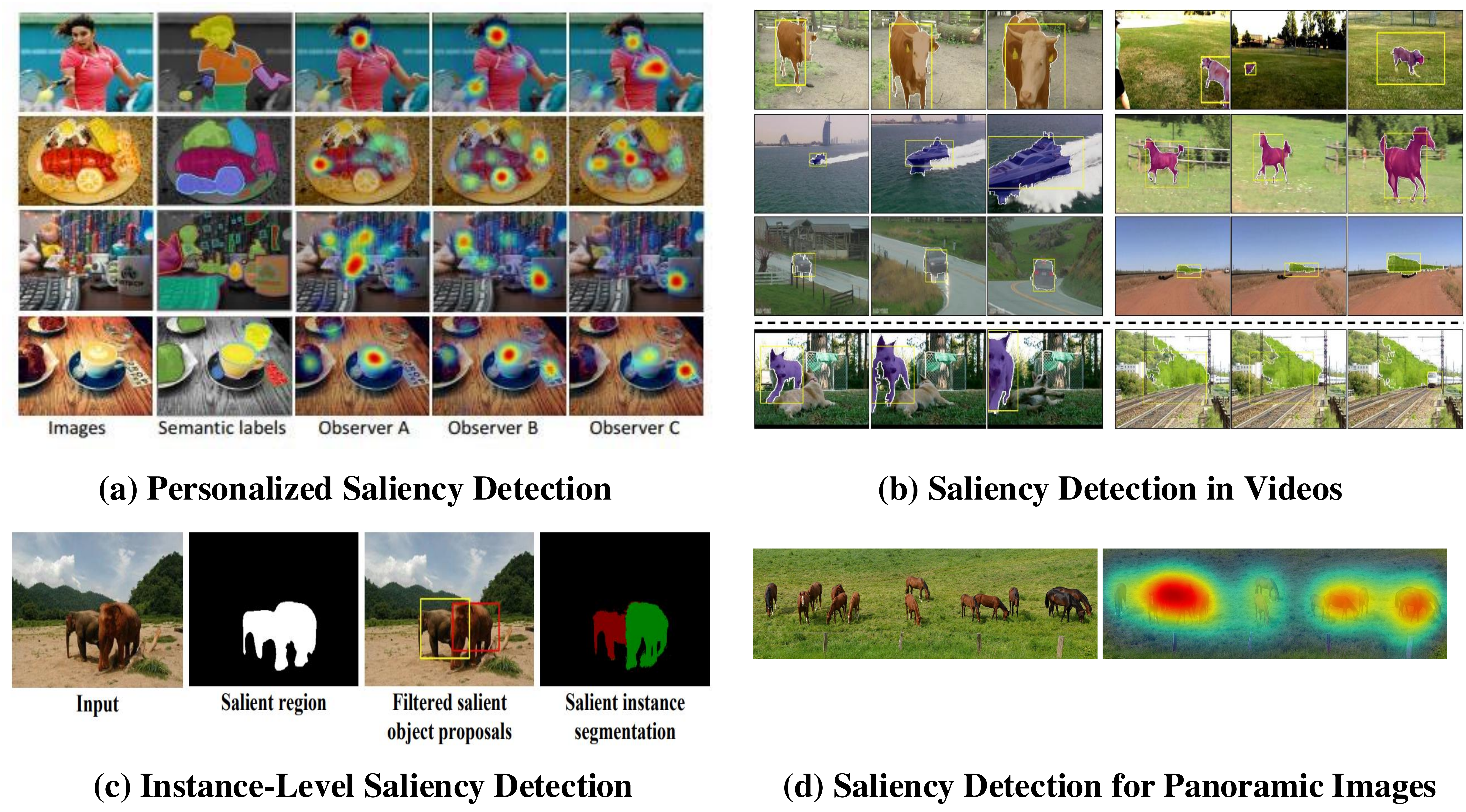}
  \caption{The examples of future directions.}
  \label{future}
\end{figure*}
In addition, we should take a look at the field of saliency detection from diversified perspectives as shown in Fig.\ref{future}. Some suggestions are given as follows:

{\bf Personalized Saliency Detection~\cite{ijcai2017-543}.} From the psychological point of view, different people have different focuses of attention (FOV) on the same scene. This is produced by different personal characteristics, such as hobbies, habits, and etc. Therefore, it is highly necessary for us to develop personalized saliency detection datasets for studying personalized saliency detection.

{\bf Saliency Detection in Videos~\cite{8066351}.} Video sequences provide temporal cues in addition to spatial information. It is difficult to explore saliency detection in videos with temporal cues, which has led to many uncertain issues such as being out of nothing, increasing or decreasing, and etc. Moreover, for saliency detection in videos, we will face all the challenges of saliency detection in images. How to incorporate temporal cues in an unified and effective way needs further study.

{\bf Instance-Level Saliency Detection~\cite{8099517}.} Existing saliency models are object-agnostic(i.e., they do not split salient regions into objects). However, similar to human capability of recognizing multiple instances of salient locations, instance-level saliency can be useful in several applications, such as image editing and video compression.

{\bf Saliency Detection for Panoramic Images~\cite{DBLP:journals/corr/abs-1710-04071}.} Almost all the previous works on saliency detection have been dedicated to conventional images. However, with the outbreak of panoramic images due to the rapid development of VR or AR technologies, it is becoming more challenging to extract salient contents in panoramic images.

\section{Conclusion}
In this paper, we proposed an innovative saliency detection algorithm using center-dark channel priors. The proposed algorithm first detects an initial saliency maps based on color and depth cues. Then we figure out the center-dark channel saliency map based on center saliency and dark channel priors. At last, we fuse them together to get the final saliency map by the updated fusion. The experiments results show that the proposed algorithm outperforms existing algorithms in both accuracy and robustness for different scenarios. Besides, by experiments, we claim that the proposed algorithm can also be applied to small object detection as well. To encourage future work, our source codes, experiment data and other related materials are all made public, which can be found on our project website\footnote {https://chunbiaozhu.github.io/ACVR2017/}.

\begin{acks}
This work was supported by the grant of National Natural Science Foundation
of China (No.U1611461), the grant of Science and Technology Planning Project of Guangdong Province,
China (No.2014B090910001) and the grant of Shenzhen Peacock Plan (No.20130408-183003656).

\end{acks}

\bibliographystyle{ACM-Reference-Format}
\bibliography{egbib}


\begin{thebibliography}{69}


\ifx \showCODEN    \undefined \def \showCODEN     #1{\unskip}     \fi
\ifx \showDOI      \undefined \def \showDOI       #1{#1}\fi
\ifx \showISBNx    \undefined \def \showISBNx     #1{\unskip}     \fi
\ifx \showISBNxiii \undefined \def \showISBNxiii  #1{\unskip}     \fi
\ifx \showISSN     \undefined \def \showISSN      #1{\unskip}     \fi
\ifx \showLCCN     \undefined \def \showLCCN      #1{\unskip}     \fi
\ifx \shownote     \undefined \def \shownote      #1{#1}          \fi
\ifx \showarticletitle \undefined \def \showarticletitle #1{#1}   \fi
\ifx \showURL      \undefined \def \showURL       {\relax}        \fi
\providecommand\bibfield[2]{#2}
\providecommand\bibinfo[2]{#2}
\providecommand\natexlab[1]{#1}
\providecommand\showeprint[2][]{arXiv:#2}

\bibitem[\protect\citeauthoryear{Achanta, Hemami, Estrada, and
  Susstrunk}{Achanta et~al\mbox{.}}{2009}]%
        {Achanta2009Frequency}
\bibfield{author}{\bibinfo{person}{Radhakrishna Achanta},
  \bibinfo{person}{Sheila Hemami}, \bibinfo{person}{Francisco Estrada}, {and}
  \bibinfo{person}{Sabine Susstrunk}.} \bibinfo{year}{2009}\natexlab{}.
\newblock \showarticletitle{Frequency-tuned salient region detection}. In
  \bibinfo{booktitle}{\emph{Computer Vision and Pattern Recognition, 2009. CVPR
  2009. IEEE Conference on}}. \bibinfo{pages}{1597--1604}.
\newblock


\bibitem[\protect\citeauthoryear{Alexe, Deselaers, and Ferrari}{Alexe
  et~al\mbox{.}}{2012}]%
        {Alexe2012Measuring}
\bibfield{author}{\bibinfo{person}{Bogdan Alexe}, \bibinfo{person}{Thomas
  Deselaers}, {and} \bibinfo{person}{Vittorio Ferrari}.}
  \bibinfo{year}{2012}\natexlab{}.
\newblock \showarticletitle{Measuring the Objectness of Image Windows}.
\newblock \bibinfo{journal}{\emph{IEEE Transactions on Pattern Analysis and
  Machine Intelligence}} \bibinfo{volume}{34}, \bibinfo{number}{11}
  (\bibinfo{year}{2012}), \bibinfo{pages}{2189--202}.
\newblock


\bibitem[\protect\citeauthoryear{Bae, Zhang, and Kweon}{Bae
  et~al\mbox{.}}{2012}]%
        {bae2012edge}
\bibfield{author}{\bibinfo{person}{Tae-Wuk Bae}, \bibinfo{person}{Fei Zhang},
  {and} \bibinfo{person}{In-So Kweon}.} \bibinfo{year}{2012}\natexlab{}.
\newblock \showarticletitle{Edge directional 2D LMS filter for infrared small
  target detection}.
\newblock \bibinfo{journal}{\emph{Infrared Physics \& Technology}}
  \bibinfo{volume}{55}, \bibinfo{number}{1} (\bibinfo{year}{2012}),
  \bibinfo{pages}{137--145}.
\newblock


\bibitem[\protect\citeauthoryear{Berman, Treibitz, and Avidan}{Berman
  et~al\mbox{.}}{2016}]%
        {Berman2016Non}
\bibfield{author}{\bibinfo{person}{Dana Berman}, \bibinfo{person}{Tali
  Treibitz}, {and} \bibinfo{person}{Shai Avidan}.}
  \bibinfo{year}{2016}\natexlab{}.
\newblock \showarticletitle{Non-local Image Dehazing}. In
  \bibinfo{booktitle}{\emph{IEEE Conference on Computer Vision and Pattern
  Recognition}}. \bibinfo{pages}{1674--1682}.
\newblock


\bibitem[\protect\citeauthoryear{Borji}{Borji}{2012}]%
        {Borji2012}
\bibfield{author}{\bibinfo{person}{A. Borji}.} \bibinfo{year}{2012}\natexlab{}.
\newblock \showarticletitle{Boosting bottom-up and top-down visual features for
  saliency estimation}. In \bibinfo{booktitle}{\emph{Computer Vision and
  Pattern Recognition}}. \bibinfo{pages}{438--445}.
\newblock


\bibitem[\protect\citeauthoryear{Borji, Cheng, Jiang, and Li}{Borji
  et~al\mbox{.}}{2014}]%
        {Borji2014Salient}
\bibfield{author}{\bibinfo{person}{Ali Borji}, \bibinfo{person}{Ming~Ming
  Cheng}, \bibinfo{person}{Huaizu Jiang}, {and} \bibinfo{person}{Jia Li}.}
  \bibinfo{year}{2014}\natexlab{}.
\newblock \showarticletitle{Salient Object Detection: A Survey}.
\newblock \bibinfo{journal}{\emph{Eprint Arxiv}} \bibinfo{volume}{16},
  \bibinfo{number}{7} (\bibinfo{year}{2014}), \bibinfo{pages}{3118}.
\newblock


\bibitem[\protect\citeauthoryear{Cerf, Harel, Einh?user, and Koch}{Cerf
  et~al\mbox{.}}{2008}]%
        {Cerf2008}
\bibfield{author}{\bibinfo{person}{Moran Cerf}, \bibinfo{person}{Jonathan
  Harel}, \bibinfo{person}{Wolfgang Einh?user}, {and} \bibinfo{person}{Christof
  Koch}.} \bibinfo{year}{2008}\natexlab{}.
\newblock \showarticletitle{Predicting human gaze using low-level saliency
  combined with face detection}. In \bibinfo{booktitle}{\emph{International
  Conference on Neural Information Processing Systems}}.
  \bibinfo{pages}{241--248}.
\newblock


\bibitem[\protect\citeauthoryear{Chang, Liang, and Chuang}{Chang
  et~al\mbox{.}}{2011}]%
        {Chang2011Content}
\bibfield{author}{\bibinfo{person}{Che~Han Chang}, \bibinfo{person}{Chia~Kai
  Liang}, {and} \bibinfo{person}{Yung~Yu Chuang}.}
  \bibinfo{year}{2011}\natexlab{}.
\newblock \showarticletitle{Content-Aware Display Adaptation and Interactive
  Editing for Stereoscopic Images}.
\newblock \bibinfo{journal}{\emph{IEEE Transactions on Multimedia}}
  \bibinfo{volume}{13}, \bibinfo{number}{4} (\bibinfo{year}{2011}),
  \bibinfo{pages}{589--601}.
\newblock


\bibitem[\protect\citeauthoryear{Chen, Li, Wei, Xia, and Tang}{Chen
  et~al\mbox{.}}{2014}]%
        {chen2014local}
\bibfield{author}{\bibinfo{person}{CL~Philip Chen}, \bibinfo{person}{Hong Li},
  \bibinfo{person}{Yantao Wei}, \bibinfo{person}{Tian Xia}, {and}
  \bibinfo{person}{Yuan~Yan Tang}.} \bibinfo{year}{2014}\natexlab{}.
\newblock \showarticletitle{A local contrast method for small infrared target
  detection}.
\newblock \bibinfo{journal}{\emph{IEEE Transactions on Geoscience and Remote
  Sensing}} \bibinfo{volume}{52}, \bibinfo{number}{1} (\bibinfo{year}{2014}),
  \bibinfo{pages}{574--581}.
\newblock


\bibitem[\protect\citeauthoryear{Cheng, Mitra, Huang, and Hu}{Cheng
  et~al\mbox{.}}{2014b}]%
        {Cheng2014SalientShape}
\bibfield{author}{\bibinfo{person}{Ming~Ming Cheng}, \bibinfo{person}{Niloy~J
  Mitra}, \bibinfo{person}{Xiaolei Huang}, {and} \bibinfo{person}{Shi~Min Hu}.}
  \bibinfo{year}{2014}\natexlab{b}.
\newblock \showarticletitle{SalientShape: group saliency in image collections}.
\newblock \bibinfo{journal}{\emph{The Visual Computer}} \bibinfo{volume}{30},
  \bibinfo{number}{4} (\bibinfo{year}{2014}), \bibinfo{pages}{443--453}.
\newblock


\bibitem[\protect\citeauthoryear{Cheng, Warrell, Lin, Zheng, Vineet, and
  Crook}{Cheng et~al\mbox{.}}{2013}]%
        {Cheng2013}
\bibfield{author}{\bibinfo{person}{Ming~Ming Cheng}, \bibinfo{person}{Jonathan
  Warrell}, \bibinfo{person}{Wen~Yan Lin}, \bibinfo{person}{Shuai Zheng},
  \bibinfo{person}{Vibhav Vineet}, {and} \bibinfo{person}{Nigel Crook}.}
  \bibinfo{year}{2013}\natexlab{}.
\newblock \showarticletitle{Efficient Salient Region Detection with Soft Image
  Abstraction}. In \bibinfo{booktitle}{\emph{IEEE International Conference on
  Computer Vision}}. \bibinfo{pages}{1529--1536}.
\newblock


\bibitem[\protect\citeauthoryear{Cheng, Fu, Wei, Xiao, and Cao}{Cheng
  et~al\mbox{.}}{2014a}]%
        {Cheng2014Depth}
\bibfield{author}{\bibinfo{person}{Yupeng Cheng}, \bibinfo{person}{Huazhu Fu},
  \bibinfo{person}{Xingxing Wei}, \bibinfo{person}{Jiangjian Xiao}, {and}
  \bibinfo{person}{Xiaochun Cao}.} \bibinfo{year}{2014}\natexlab{a}.
\newblock \showarticletitle{Depth Enhanced Saliency Detection Method}.
\newblock  \bibinfo{volume}{55}, \bibinfo{number}{1} (\bibinfo{year}{2014}),
  \bibinfo{pages}{23--27}.
\newblock


\bibitem[\protect\citeauthoryear{Furnari, Farinella, and Battiato}{Furnari
  et~al\mbox{.}}{2014}]%
        {Furnari2014An}
\bibfield{author}{\bibinfo{person}{Antonino Furnari},
  \bibinfo{person}{Giovanni~Maria Farinella}, {and} \bibinfo{person}{Sebastiano
  Battiato}.} \bibinfo{year}{2014}\natexlab{}.
\newblock \showarticletitle{An Experimental Analysis of Saliency Detection with
  Respect to Three Saliency Levels}. In \bibinfo{booktitle}{\emph{Workshop at
  the European Conference on Computer Vision}}. \bibinfo{pages}{806--821}.
\newblock


\bibitem[\protect\citeauthoryear{Gao, Han, and Vasconcelos}{Gao
  et~al\mbox{.}}{2009}]%
        {Gao2009}
\bibfield{author}{\bibinfo{person}{Dashan Gao}, \bibinfo{person}{Sunhyoung
  Han}, {and} \bibinfo{person}{Nuno Vasconcelos}.}
  \bibinfo{year}{2009}\natexlab{}.
\newblock \showarticletitle{Discriminant Saliency, the Detection of Suspicious
  Coincidences, and Applications to Visual Recognition}.
\newblock \bibinfo{journal}{\emph{IEEE Transactions on Pattern Analysis and
  Machine Intelligence}} \bibinfo{volume}{31}, \bibinfo{number}{6}
  (\bibinfo{year}{2009}), \bibinfo{pages}{989--1005}.
\newblock


\bibitem[\protect\citeauthoryear{Gao and Vasconcelos}{Gao and
  Vasconcelos}{2004}]%
        {Gao2005}
\bibfield{author}{\bibinfo{person}{Dashan Gao} {and} \bibinfo{person}{Nuno
  Vasconcelos}.} \bibinfo{year}{2004}\natexlab{}.
\newblock \showarticletitle{Discriminant Saliency for Visual Recognition from
  Cluttered Scenes}.
\newblock \bibinfo{journal}{\emph{Advances in Neural Information Processing
  Systems}}  \bibinfo{volume}{17} (\bibinfo{year}{2004}),
  \bibinfo{pages}{481--488}.
\newblock


\bibitem[\protect\citeauthoryear{Geng}{Geng}{2012}]%
        {Geng2012Leveraging}
\bibfield{author}{\bibinfo{person}{Yujie Geng}.}
  \bibinfo{year}{2012}\natexlab{}.
\newblock \showarticletitle{Leveraging stereopsis for saliency analysis}. In
  \bibinfo{booktitle}{\emph{Computer Vision and Pattern Recognition}}.
  \bibinfo{pages}{454--461}.
\newblock


\bibitem[\protect\citeauthoryear{Girshick, Donahue, Darrell, and
  Malik}{Girshick et~al\mbox{.}}{2013}]%
        {Girshick2013Rich}
\bibfield{author}{\bibinfo{person}{Ross Girshick}, \bibinfo{person}{Jeff
  Donahue}, \bibinfo{person}{Trevor Darrell}, {and} \bibinfo{person}{Jitendra
  Malik}.} \bibinfo{year}{2013}\natexlab{}.
\newblock \showarticletitle{Rich Feature Hierarchies for Accurate Object
  Detection and Semantic Segmentation}. In \bibinfo{booktitle}{\emph{IEEE
  Conference on Computer Vision and Pattern Recognition}}.
  \bibinfo{pages}{580--587}.
\newblock


\bibitem[\protect\citeauthoryear{He, Sun, and Tang}{He et~al\mbox{.}}{2009}]%
        {He2009Single}
\bibfield{author}{\bibinfo{person}{Kaiming He}, \bibinfo{person}{Jian Sun},
  {and} \bibinfo{person}{Xiaoou Tang}.} \bibinfo{year}{2009}\natexlab{}.
\newblock \showarticletitle{Single image haze removal using dark channel
  prior}. In \bibinfo{booktitle}{\emph{Computer Vision and Pattern Recognition,
  2009. CVPR 2009. IEEE Conference on}}. \bibinfo{pages}{1956--1963}.
\newblock


\bibitem[\protect\citeauthoryear{He, Zhang, Ren, and Sun}{He
  et~al\mbox{.}}{2016}]%
        {He2016Delving}
\bibfield{author}{\bibinfo{person}{Kaiming He}, \bibinfo{person}{Xiangyu
  Zhang}, \bibinfo{person}{Shaoqing Ren}, {and} \bibinfo{person}{Jian Sun}.}
  \bibinfo{year}{2016}\natexlab{}.
\newblock \showarticletitle{Delving Deep into Rectifiers: Surpassing
  Human-Level Performance on ImageNet Classification}. In
  \bibinfo{booktitle}{\emph{IEEE International Conference on Computer Vision}}.
  \bibinfo{pages}{1026--1034}.
\newblock


\bibitem[\protect\citeauthoryear{Hornung, Pritch, Krahenbuhl, and
  Perazzi}{Hornung et~al\mbox{.}}{2012}]%
        {Perazzi2012}
\bibfield{author}{\bibinfo{person}{A. Hornung}, \bibinfo{person}{Y. Pritch},
  \bibinfo{person}{P. Krahenbuhl}, {and} \bibinfo{person}{F. Perazzi}.}
  \bibinfo{year}{2012}\natexlab{}.
\newblock \showarticletitle{Saliency filters: Contrast based filtering for
  salient region detection}. In \bibinfo{booktitle}{\emph{IEEE Conference on
  Computer Vision and Pattern Recognition}}. \bibinfo{pages}{733--740}.
\newblock


\bibitem[\protect\citeauthoryear{Itti, Koch, and Niebur}{Itti
  et~al\mbox{.}}{1998}]%
        {Itti1998A}
\bibfield{author}{\bibinfo{person}{L Itti}, \bibinfo{person}{C Koch}, {and}
  \bibinfo{person}{E Niebur}.} \bibinfo{year}{1998}\natexlab{}.
\newblock \showarticletitle{A Model of Saliency-Based Visual Attention for
  Rapid Scene Analysis}.
\newblock \bibinfo{journal}{\emph{IEEE Transactions on Pattern Analysis and
  Machine Intelligence}} \bibinfo{volume}{20}, \bibinfo{number}{11}
  (\bibinfo{year}{1998}), \bibinfo{pages}{1254--1259}.
\newblock


\bibitem[\protect\citeauthoryear{Ji, Yao, Tian, Xu, Sun, and Liu}{Ji
  et~al\mbox{.}}{2012}]%
        {Ji:2012:CSF:2168752.2168758}
\bibfield{author}{\bibinfo{person}{Rongrong Ji}, \bibinfo{person}{Hongxun Yao},
  \bibinfo{person}{Qi Tian}, \bibinfo{person}{Pengfei Xu},
  \bibinfo{person}{Xiaoshuai Sun}, {and} \bibinfo{person}{Xianming Liu}.}
  \bibinfo{year}{2012}\natexlab{}.
\newblock \showarticletitle{Context-Aware Semi-Local Feature Detector}.
\newblock \bibinfo{journal}{\emph{ACM Trans. Intell. Syst. Technol.}}
  \bibinfo{volume}{3}, \bibinfo{number}{3}, Article \bibinfo{articleno}{44}
  (\bibinfo{date}{May} \bibinfo{year}{2012}), \bibinfo{numpages}{27}~pages.
\newblock
\showISSN{2157-6904}
\urldef\tempurl%
\url{https://doi.org/10.1145/2168752.2168758}
\showDOI{\tempurl}


\bibitem[\protect\citeauthoryear{Jr}{Jr}{1988}]%
        {Jr1988An}
\bibfield{author}{\bibinfo{person}{Pat S.~Chavez Jr}.}
  \bibinfo{year}{1988}\natexlab{}.
\newblock \showarticletitle{An improved dark-object subtraction technique for
  atmospheric scattering correction of multispectral data ☆}.
\newblock \bibinfo{journal}{\emph{Remote Sensing of Environment}}
  \bibinfo{volume}{24}, \bibinfo{number}{3} (\bibinfo{year}{1988}),
  \bibinfo{pages}{459--479}.
\newblock


\bibitem[\protect\citeauthoryear{Ju, Ge, Geng, Ren, and Wu}{Ju
  et~al\mbox{.}}{2015}]%
        {Ju2015Depth}
\bibfield{author}{\bibinfo{person}{Ran Ju}, \bibinfo{person}{Ling Ge},
  \bibinfo{person}{Wenjing Geng}, \bibinfo{person}{Tongwei Ren}, {and}
  \bibinfo{person}{Gangshan Wu}.} \bibinfo{year}{2015}\natexlab{}.
\newblock \showarticletitle{Depth saliency based on anisotropic center-surround
  difference}. In \bibinfo{booktitle}{\emph{ICIP}}.
  \bibinfo{pages}{1115--1119}.
\newblock


\bibitem[\protect\citeauthoryear{Judd, Ehinger, Durand, and Torralba}{Judd
  et~al\mbox{.}}{2009}]%
        {Judd2009Learning}
\bibfield{author}{\bibinfo{person}{T Judd}, \bibinfo{person}{K Ehinger},
  \bibinfo{person}{F Durand}, {and} \bibinfo{person}{A Torralba}.}
  \bibinfo{year}{2009}\natexlab{}.
\newblock \showarticletitle{Learning to predict where humans look}. In
  \bibinfo{booktitle}{\emph{IEEE International Conference on Computer Vision,
  ICCV 2009, Kyoto, Japan, September 27 - October}}.
  \bibinfo{pages}{2106--2113}.
\newblock


\bibitem[\protect\citeauthoryear{Kim, Jang, Sim, and Kim}{Kim
  et~al\mbox{.}}{2013}]%
        {kim2013optimized}
\bibfield{author}{\bibinfo{person}{Jin-Hwan Kim}, \bibinfo{person}{Won-Dong
  Jang}, \bibinfo{person}{Jae-Young Sim}, {and} \bibinfo{person}{Chang-Su
  Kim}.} \bibinfo{year}{2013}\natexlab{}.
\newblock \showarticletitle{Optimized contrast enhancement for real-time image
  and video dehazing}.
\newblock \bibinfo{journal}{\emph{Journal of Visual Communication and Image
  Representation}} \bibinfo{volume}{24}, \bibinfo{number}{3}
  (\bibinfo{year}{2013}), \bibinfo{pages}{410--425}.
\newblock


\bibitem[\protect\citeauthoryear{Kim, Sim, and Kim}{Kim et~al\mbox{.}}{2014}]%
        {Kim2014}
\bibfield{author}{\bibinfo{person}{Jun~Seong Kim}, \bibinfo{person}{Jae~Young
  Sim}, {and} \bibinfo{person}{Chang~Su Kim}.} \bibinfo{year}{2014}\natexlab{}.
\newblock \showarticletitle{Multiscale Saliency Detection Using Random Walk
  With Restart}.
\newblock \bibinfo{journal}{\emph{IEEE Transactions on Circuits and Systems for
  Video Technology}} \bibinfo{volume}{24}, \bibinfo{number}{2}
  (\bibinfo{year}{2014}), \bibinfo{pages}{198--210}.
\newblock


\bibitem[\protect\citeauthoryear{Lee, Gu, and Park}{Lee et~al\mbox{.}}{2015}]%
        {lee2015effective}
\bibfield{author}{\bibinfo{person}{Eunyoung Lee}, \bibinfo{person}{Eunhye Gu},
  {and} \bibinfo{person}{Kilhoum Park}.} \bibinfo{year}{2015}\natexlab{}.
\newblock \showarticletitle{Effective small target enhancement and detection in
  infrared images using saliency map and image intensity}.
\newblock \bibinfo{journal}{\emph{Optical Review}} \bibinfo{volume}{22},
  \bibinfo{number}{4} (\bibinfo{year}{2015}), \bibinfo{pages}{659--668}.
\newblock


\bibitem[\protect\citeauthoryear{Leo, Medioni, Trivedi, Kanade, and
  Farinella}{Leo et~al\mbox{.}}{2017}]%
        {leo2017computer}
\bibfield{author}{\bibinfo{person}{Marco Leo}, \bibinfo{person}{G Medioni},
  \bibinfo{person}{M Trivedi}, \bibinfo{person}{Takeo Kanade}, {and}
  \bibinfo{person}{Giovanni~Maria Farinella}.} \bibinfo{year}{2017}\natexlab{}.
\newblock \showarticletitle{Computer vision for assistive technologies}.
\newblock \bibinfo{journal}{\emph{Computer Vision and Image Understanding}}
  \bibinfo{volume}{154} (\bibinfo{year}{2017}), \bibinfo{pages}{1--15}.
\newblock


\bibitem[\protect\citeauthoryear{Li, Xie, Lin, and Yu}{Li
  et~al\mbox{.}}{2017}]%
        {8099517}
\bibfield{author}{\bibinfo{person}{G. Li}, \bibinfo{person}{Y. Xie},
  \bibinfo{person}{L. Lin}, {and} \bibinfo{person}{Y. Yu}.}
  \bibinfo{year}{2017}\natexlab{}.
\newblock \showarticletitle{Instance-Level Salient Object Segmentation}. In
  \bibinfo{booktitle}{\emph{2017 IEEE Conference on Computer Vision and Pattern
  Recognition (CVPR)}}. \bibinfo{pages}{247--256}.
\newblock
\showISSN{1063-6919}
\urldef\tempurl%
\url{https://doi.org/10.1109/CVPR.2017.34}
\showDOI{\tempurl}


\bibitem[\protect\citeauthoryear{Li and Yu}{Li and Yu}{2015}]%
        {Li2015Visual}
\bibfield{author}{\bibinfo{person}{Guanbin Li} {and} \bibinfo{person}{Yizhou
  Yu}.} \bibinfo{year}{2015}\natexlab{}.
\newblock \showarticletitle{Visual saliency based on multiscale deep features}.
  In \bibinfo{booktitle}{\emph{Computer Vision and Pattern Recognition}}.
  \bibinfo{pages}{5455--5463}.
\newblock


\bibitem[\protect\citeauthoryear{Li and Yu}{Li and Yu}{2016}]%
        {Li2016Deep}
\bibfield{author}{\bibinfo{person}{Guanbin Li} {and} \bibinfo{person}{Yizhou
  Yu}.} \bibinfo{year}{2016}\natexlab{}.
\newblock \showarticletitle{Deep Contrast Learning for Salient Object
  Detection}. In \bibinfo{booktitle}{\emph{Computer Vision and Pattern
  Recognition}}. \bibinfo{pages}{478--487}.
\newblock


\bibitem[\protect\citeauthoryear{Li and Zhu}{Li and Zhu}{2017}]%
        {8265566}
\bibfield{author}{\bibinfo{person}{G. Li} {and} \bibinfo{person}{C. Zhu}.}
  \bibinfo{year}{2017}\natexlab{}.
\newblock \showarticletitle{A Three-Pathway Psychobiological Framework of
  Salient Object Detection Using Stereoscopic Technology}. In
  \bibinfo{booktitle}{\emph{2017 ICCVW}}. \bibinfo{pages}{3008--3014}.
\newblock
\urldef\tempurl%
\url{https://doi.org/10.1109/ICCVW.2017.355}
\showDOI{\tempurl}


\bibitem[\protect\citeauthoryear{Li, Lu, Lin, Shen, and Price}{Li
  et~al\mbox{.}}{2015a}]%
        {Li2015Inner}
\bibfield{author}{\bibinfo{person}{Hongyang Li}, \bibinfo{person}{Huchuan Lu},
  \bibinfo{person}{Zhe Lin}, \bibinfo{person}{Xiaohui Shen}, {and}
  \bibinfo{person}{Brian Price}.} \bibinfo{year}{2015}\natexlab{a}.
\newblock \showarticletitle{Inner and Inter Label Propagation: Salient Object
  Detection in the Wild}.
\newblock \bibinfo{journal}{\emph{IEEE Transactions on Image Processing A
  Publication of the IEEE Signal Processing Society}} \bibinfo{volume}{24},
  \bibinfo{number}{10} (\bibinfo{year}{2015}), \bibinfo{pages}{3176}.
\newblock


\bibitem[\protect\citeauthoryear{Li, Lu, Lin, Shen, and Price}{Li
  et~al\mbox{.}}{2015b}]%
        {Li2015}
\bibfield{author}{\bibinfo{person}{Hongyang Li}, \bibinfo{person}{Huchuan Lu},
  \bibinfo{person}{Zhe Lin}, \bibinfo{person}{Xiaohui Shen}, {and}
  \bibinfo{person}{Brian Price}.} \bibinfo{year}{2015}\natexlab{b}.
\newblock \bibinfo{booktitle}{\emph{Inner and Inter Label Propagation: Salient
  Object Detection in the Wild}}.
\newblock \bibinfo{publisher}{New Park Pub.,}. 3176--86 pages.
\newblock


\bibitem[\protect\citeauthoryear{Li, Xia, and Chen}{Li et~al\mbox{.}}{2018}]%
        {8066351}
\bibfield{author}{\bibinfo{person}{J. Li}, \bibinfo{person}{C. Xia}, {and}
  \bibinfo{person}{X. Chen}.} \bibinfo{year}{2018}\natexlab{}.
\newblock \showarticletitle{A Benchmark Dataset and Saliency-Guided Stacked
  Autoencoders for Video-Based Salient Object Detection}.
\newblock \bibinfo{journal}{\emph{IEEE Transactions on Image Processing}}
  \bibinfo{volume}{27}, \bibinfo{number}{1} (\bibinfo{date}{Jan}
  \bibinfo{year}{2018}), \bibinfo{pages}{349--364}.
\newblock
\showISSN{1057-7149}
\urldef\tempurl%
\url{https://doi.org/10.1109/TIP.2017.2762594}
\showDOI{\tempurl}


\bibitem[\protect\citeauthoryear{Li, Pan, and Liu}{Li et~al\mbox{.}}{2009}]%
        {li2009saliency}
\bibfield{author}{\bibinfo{person}{Wei Li}, \bibinfo{person}{Chunhong Pan},
  {and} \bibinfo{person}{Li-xiong Liu}.} \bibinfo{year}{2009}\natexlab{}.
\newblock \showarticletitle{Saliency-based automatic target detection in
  forward looking infrared images}. In \bibinfo{booktitle}{\emph{Image
  Processing (ICIP), 2009 16th IEEE International Conference on}}. IEEE,
  \bibinfo{pages}{957--960}.
\newblock


\bibitem[\protect\citeauthoryear{Li, Hou, Koch, Rehg, and Yuille}{Li
  et~al\mbox{.}}{2014}]%
        {Li2014The}
\bibfield{author}{\bibinfo{person}{Yin Li}, \bibinfo{person}{Xiaodi Hou},
  \bibinfo{person}{Christof Koch}, \bibinfo{person}{James~M. Rehg}, {and}
  \bibinfo{person}{Alan~L. Yuille}.} \bibinfo{year}{2014}\natexlab{}.
\newblock \showarticletitle{The Secrets of Salient Object Segmentation}. In
  \bibinfo{booktitle}{\emph{IEEE Conference on Computer Vision and Pattern
  Recognition}}. \bibinfo{pages}{280--287}.
\newblock


\bibitem[\protect\citeauthoryear{Lou, Zhu, Wang, and Ren}{Lou
  et~al\mbox{.}}{2016}]%
        {Lou2016Small}
\bibfield{author}{\bibinfo{person}{Jing Lou}, \bibinfo{person}{Wei Zhu},
  \bibinfo{person}{Huan Wang}, {and} \bibinfo{person}{Mingwu Ren}.}
  \bibinfo{year}{2016}\natexlab{}.
\newblock \showarticletitle{Small target detection combining regional stability
  and saliency in a color image}.
\newblock \bibinfo{journal}{\emph{Multimedia Tools and Applications}}
  (\bibinfo{year}{2016}), \bibinfo{pages}{1--18}.
\newblock


\bibitem[\protect\citeauthoryear{Lu, Zhang, Qi, Na, Xiang, and Yang}{Lu
  et~al\mbox{.}}{2016}]%
        {Lu2016Co}
\bibfield{author}{\bibinfo{person}{Huchuan Lu}, \bibinfo{person}{Xiaoning
  Zhang}, \bibinfo{person}{Jinqing Qi}, \bibinfo{person}{Tong Na},
  \bibinfo{person}{Ruan Xiang}, {and} \bibinfo{person}{Ming~Hsuan Yang}.}
  \bibinfo{year}{2016}\natexlab{}.
\newblock \showarticletitle{Co-bootstrapping Saliency}.
\newblock \bibinfo{journal}{\emph{IEEE Transactions on Image Processing A
  Publication of the IEEE Signal Processing Society}} \bibinfo{volume}{PP},
  \bibinfo{number}{99} (\bibinfo{year}{2016}), \bibinfo{pages}{1--1}.
\newblock


\bibitem[\protect\citeauthoryear{Luo, Yuan, Xue, and Tian}{Luo
  et~al\mbox{.}}{2011}]%
        {Luo2011}
\bibfield{author}{\bibinfo{person}{Ye Luo}, \bibinfo{person}{Junsong Yuan},
  \bibinfo{person}{Ping Xue}, {and} \bibinfo{person}{Qi Tian}.}
  \bibinfo{year}{2011}\natexlab{}.
\newblock \showarticletitle{Saliency Density Maximization for Efficient Visual
  Objects Discovery}.
\newblock \bibinfo{journal}{\emph{IEEE Transactions on Circuits and Systems for
  Video Technology}} \bibinfo{volume}{21}, \bibinfo{number}{12}
  (\bibinfo{year}{2011}), \bibinfo{pages}{1822--1834}.
\newblock


\bibitem[\protect\citeauthoryear{Murray, Vanrell, Otazu, and Parraga}{Murray
  et~al\mbox{.}}{2011}]%
        {Murray2011Saliency}
\bibfield{author}{\bibinfo{person}{N. Murray}, \bibinfo{person}{M. Vanrell},
  \bibinfo{person}{X. Otazu}, {and} \bibinfo{person}{C.~A. Parraga}.}
  \bibinfo{year}{2011}\natexlab{}.
\newblock \showarticletitle{Saliency estimation using a non-parametric
  low-level vision model}.
\newblock  \bibinfo{volume}{42}, \bibinfo{number}{7} (\bibinfo{year}{2011}),
  \bibinfo{pages}{433--440}.
\newblock


\bibitem[\protect\citeauthoryear{Oliva, Torralba, Castelhano, and
  Henderson}{Oliva et~al\mbox{.}}{2003}]%
        {Oliva2003}
\bibfield{author}{\bibinfo{person}{A. Oliva}, \bibinfo{person}{A. Torralba},
  \bibinfo{person}{M.~S. Castelhano}, {and} \bibinfo{person}{J.~M. Henderson}.}
  \bibinfo{year}{2003}\natexlab{}.
\newblock \showarticletitle{Top-down control of visual attention in object
  detection}. In \bibinfo{booktitle}{\emph{International Conference on Image
  Processing, 2003. ICIP 2003. Proceedings}}. \bibinfo{pages}{I--253--6 vol.1}.
\newblock


\bibitem[\protect\citeauthoryear{Peng, Li, Xiong, Hu, and Ji}{Peng
  et~al\mbox{.}}{2014}]%
        {Peng2014RGBD}
\bibfield{author}{\bibinfo{person}{Houwen Peng}, \bibinfo{person}{Bing Li},
  \bibinfo{person}{Weihua Xiong}, \bibinfo{person}{Weiming Hu}, {and}
  \bibinfo{person}{Rongrong Ji}.} \bibinfo{year}{2014}\natexlab{}.
\newblock \bibinfo{booktitle}{\emph{RGBD Salient Object Detection: A Benchmark
  and Algorithms}}.
\newblock \bibinfo{publisher}{Springer International Publishing}. 92--109
  pages.
\newblock


\bibitem[\protect\citeauthoryear{Qi, Ma, Tao, Yang, and Tian}{Qi
  et~al\mbox{.}}{2013}]%
        {qi2013robust}
\bibfield{author}{\bibinfo{person}{Shengxiang Qi}, \bibinfo{person}{Jie Ma},
  \bibinfo{person}{Chao Tao}, \bibinfo{person}{Changcai Yang}, {and}
  \bibinfo{person}{Jinwen Tian}.} \bibinfo{year}{2013}\natexlab{}.
\newblock \showarticletitle{A robust directional saliency-based method for
  infrared small-target detection under various complex backgrounds}.
\newblock \bibinfo{journal}{\emph{IEEE Geoscience and Remote Sensing Letters}}
  \bibinfo{volume}{10}, \bibinfo{number}{3} (\bibinfo{year}{2013}),
  \bibinfo{pages}{495--499}.
\newblock


\bibitem[\protect\citeauthoryear{Qin, Lu, Xu, and Wang}{Qin
  et~al\mbox{.}}{2015}]%
        {Qin2015Saliency}
\bibfield{author}{\bibinfo{person}{Yao Qin}, \bibinfo{person}{Huchuan Lu},
  \bibinfo{person}{Yiqun Xu}, {and} \bibinfo{person}{He Wang}.}
  \bibinfo{year}{2015}\natexlab{}.
\newblock \showarticletitle{Saliency detection via Cellular Automata}. In
  \bibinfo{booktitle}{\emph{Computer Vision and Pattern Recognition}}.
  \bibinfo{pages}{110--119}.
\newblock


\bibitem[\protect\citeauthoryear{Ran, Tal, and Zelnikmanor}{Ran
  et~al\mbox{.}}{2013}]%
        {Margolin2013}
\bibfield{author}{\bibinfo{person}{Margolin Ran}, \bibinfo{person}{Ayellet
  Tal}, {and} \bibinfo{person}{Lihi Zelnikmanor}.}
  \bibinfo{year}{2013}\natexlab{}.
\newblock \showarticletitle{What Makes a Patch Distinct?}. In
  \bibinfo{booktitle}{\emph{IEEE Conference on Computer Vision and Pattern
  Recognition}}. \bibinfo{pages}{1139--1146}.
\newblock


\bibitem[\protect\citeauthoryear{Rao, Huang, and Fu}{Rao et~al\mbox{.}}{2016}]%
        {Rao:2016:LHC:2906145.2873065}
\bibfield{author}{\bibinfo{person}{Huaming Rao}, \bibinfo{person}{Shih-Wen
  Huang}, {and} \bibinfo{person}{Wai-Tat Fu}.} \bibinfo{year}{2016}\natexlab{}.
\newblock \showarticletitle{Leveraging Human Computations to Improve
  Schematization of Spatial Relations from Imagery}.
\newblock \bibinfo{journal}{\emph{ACM Trans. Intell. Syst. Technol.}}
  \bibinfo{volume}{7}, \bibinfo{number}{4}, Article \bibinfo{articleno}{54}
  (\bibinfo{date}{March} \bibinfo{year}{2016}), \bibinfo{numpages}{21}~pages.
\newblock
\showISSN{2157-6904}
\urldef\tempurl%
\url{https://doi.org/10.1145/2873065}
\showDOI{\tempurl}


\bibitem[\protect\citeauthoryear{Shi, Yan, Li, and Jia}{Shi
  et~al\mbox{.}}{2016a}]%
        {Shi2016Hierarchical}
\bibfield{author}{\bibinfo{person}{Jianping Shi}, \bibinfo{person}{Qiong Yan},
  \bibinfo{person}{Xu Li}, {and} \bibinfo{person}{Jiaya Jia}.}
  \bibinfo{year}{2016}\natexlab{a}.
\newblock \showarticletitle{Hierarchical Image Saliency Detection on Extended
  CSSD}.
\newblock \bibinfo{journal}{\emph{IEEE Transactions on Pattern Analysis and
  Machine Intelligence}} \bibinfo{volume}{38}, \bibinfo{number}{4}
  (\bibinfo{year}{2016}), \bibinfo{pages}{717--729}.
\newblock


\bibitem[\protect\citeauthoryear{Shi, Yan, Li, and Jia}{Shi
  et~al\mbox{.}}{2016b}]%
        {Shi2016}
\bibfield{author}{\bibinfo{person}{Jianping Shi}, \bibinfo{person}{Qiong Yan},
  \bibinfo{person}{Xu Li}, {and} \bibinfo{person}{Jiaya Jia}.}
  \bibinfo{year}{2016}\natexlab{b}.
\newblock \showarticletitle{Hierarchical Image Saliency Detection on Extended
  CSSD}.
\newblock \bibinfo{journal}{\emph{IEEE Transactions on Pattern Analysis and
  Machine Intelligence}} \bibinfo{volume}{38}, \bibinfo{number}{4}
  (\bibinfo{year}{2016}), \bibinfo{pages}{717}.
\newblock


\bibitem[\protect\citeauthoryear{Sun and Ling}{Sun and Ling}{2011}]%
        {Sun2011Scale}
\bibfield{author}{\bibinfo{person}{Jin Sun} {and} \bibinfo{person}{Haibin
  Ling}.} \bibinfo{year}{2011}\natexlab{}.
\newblock \showarticletitle{Scale and object aware image retargeting for
  thumbnail browsing}. In \bibinfo{booktitle}{\emph{International Conference on
  Computer Vision}}. \bibinfo{pages}{1511--1518}.
\newblock


\bibitem[\protect\citeauthoryear{Tilke, Ehinger, Durand, and Torralba}{Tilke
  et~al\mbox{.}}{2009}]%
        {Judd2009}
\bibfield{author}{\bibinfo{person}{Juddk Tilke}, \bibinfo{person}{Krista
  Ehinger}, \bibinfo{person}{Frédo Durand}, {and} \bibinfo{person}{Antonio
  Torralba}.} \bibinfo{year}{2009}\natexlab{}.
\newblock \showarticletitle{Learning to predict where humans look}.
\newblock  \bibinfo{volume}{30}, \bibinfo{number}{2} (\bibinfo{year}{2009}),
  \bibinfo{pages}{2106--2113}.
\newblock


\bibitem[\protect\citeauthoryear{Tong, Lu, Xiang, and Yang}{Tong
  et~al\mbox{.}}{2015}]%
        {Tong2015Salient}
\bibfield{author}{\bibinfo{person}{Na Tong}, \bibinfo{person}{Huchuan Lu},
  \bibinfo{person}{Ruan Xiang}, {and} \bibinfo{person}{Ming~Hsuan Yang}.}
  \bibinfo{year}{2015}\natexlab{}.
\newblock \showarticletitle{Salient object detection via bootstrap learning}.
  In \bibinfo{booktitle}{\emph{Computer Vision and Pattern Recognition}}.
  \bibinfo{pages}{1884--1892}.
\newblock


\bibitem[\protect\citeauthoryear{Wang, Lu, Wang, Feng, Wang, Yin, and
  Ruan}{Wang et~al\mbox{.}}{2017}]%
        {wang2017learning}
\bibfield{author}{\bibinfo{person}{Lijun Wang}, \bibinfo{person}{Huchuan Lu},
  \bibinfo{person}{Yifan Wang}, \bibinfo{person}{Mengyang Feng},
  \bibinfo{person}{Dong Wang}, \bibinfo{person}{Baocai Yin}, {and}
  \bibinfo{person}{Xiang Ruan}.} \bibinfo{year}{2017}\natexlab{}.
\newblock \showarticletitle{Learning to Detect Salient Objects With Image-Level
  Supervision}. In \bibinfo{booktitle}{\emph{Proceedings of the IEEE Conference
  on Computer Vision and Pattern Recognition}}. \bibinfo{pages}{136--145}.
\newblock


\bibitem[\protect\citeauthoryear{Wang, Wang, Lu, Zhang, and Xiang}{Wang
  et~al\mbox{.}}{2016}]%
        {Wang2016Saliency}
\bibfield{author}{\bibinfo{person}{Linzhao Wang}, \bibinfo{person}{Lijun Wang},
  \bibinfo{person}{Huchuan Lu}, \bibinfo{person}{Pingping Zhang}, {and}
  \bibinfo{person}{Ruan Xiang}.} \bibinfo{year}{2016}\natexlab{}.
\newblock \showarticletitle{Saliency Detection with Recurrent Fully
  Convolutional Networks}. In \bibinfo{booktitle}{\emph{European Conference on
  Computer Vision}}. \bibinfo{pages}{825--841}.
\newblock


\bibitem[\protect\citeauthoryear{Xie, Guo, and Cai}{Xie et~al\mbox{.}}{2010}]%
        {Xie2010Improved}
\bibfield{author}{\bibinfo{person}{Bin Xie}, \bibinfo{person}{Fan Guo}, {and}
  \bibinfo{person}{Zixing Cai}.} \bibinfo{year}{2010}\natexlab{}.
\newblock \showarticletitle{Improved Single Image Dehazing Using Dark Channel
  Prior and Multi-scale Retinex}. In \bibinfo{booktitle}{\emph{International
  Conference on Intelligent System Design and Engineering Application}}.
  \bibinfo{pages}{848--851}.
\newblock


\bibitem[\protect\citeauthoryear{Yang, Zhang, and Lu}{Yang
  et~al\mbox{.}}{2013}]%
        {Yang2013}
\bibfield{author}{\bibinfo{person}{Chuan Yang}, \bibinfo{person}{Lihe Zhang},
  {and} \bibinfo{person}{Huchuan Lu}.} \bibinfo{year}{2013}\natexlab{}.
\newblock \showarticletitle{Graph-Regularized Saliency Detection With
  Convex-Hull-Based Center Prior}.
\newblock \bibinfo{journal}{\emph{IEEE Signal Processing Letters}}
  \bibinfo{volume}{20}, \bibinfo{number}{7} (\bibinfo{year}{2013}),
  \bibinfo{pages}{637--640}.
\newblock


\bibitem[\protect\citeauthoryear{Yang}{Yang}{2012}]%
        {Yang2012}
\bibfield{author}{\bibinfo{person}{Jimei Yang}.}
  \bibinfo{year}{2012}\natexlab{}.
\newblock \showarticletitle{Top-down visual saliency via joint CRF and
  dictionary learning}. In \bibinfo{booktitle}{\emph{Computer Vision and
  Pattern Recognition}}. \bibinfo{pages}{2296--2303}.
\newblock


\bibitem[\protect\citeauthoryear{Yanyu~Xu}{Yanyu~Xu}{2017}]%
        {ijcai2017-543}
\bibfield{author}{\bibinfo{person}{Junru Wu Jingyi Yu Shenghua~Gao Yanyu~Xu,
  Nianyi~Li}.} \bibinfo{year}{2017}\natexlab{}.
\newblock \showarticletitle{Beyond Universal Saliency: Personalized Saliency
  Prediction with Multi-task CNN}. In \bibinfo{booktitle}{\emph{Proceedings of
  the Twenty-Sixth International Joint Conference on Artificial Intelligence,
  {IJCAI-17}}}. \bibinfo{pages}{3887--3893}.
\newblock
\urldef\tempurl%
\url{https://doi.org/10.24963/ijcai.2017/543}
\showDOI{\tempurl}


\bibitem[\protect\citeauthoryear{Zhang, Fu, Han, Borji, and Li}{Zhang
  et~al\mbox{.}}{2018}]%
        {Zhang:2018:RCD:3183892.3158674}
\bibfield{author}{\bibinfo{person}{Dingwen Zhang}, \bibinfo{person}{Huazhu Fu},
  \bibinfo{person}{Junwei Han}, \bibinfo{person}{Ali Borji}, {and}
  \bibinfo{person}{Xuelong Li}.} \bibinfo{year}{2018}\natexlab{}.
\newblock \showarticletitle{A Review of Co-Saliency Detection Algorithms:
  Fundamentals, Applications, and Challenges}.
\newblock \bibinfo{journal}{\emph{ACM Trans. Intell. Syst. Technol.}}
  \bibinfo{volume}{9}, \bibinfo{number}{4}, Article \bibinfo{articleno}{38}
  (\bibinfo{date}{Jan.} \bibinfo{year}{2018}), \bibinfo{numpages}{31}~pages.
\newblock
\showISSN{2157-6904}
\urldef\tempurl%
\url{https://doi.org/10.1145/3158674}
\showDOI{\tempurl}


\bibitem[\protect\citeauthoryear{Zhang, Wang, Lu, Wang, and Ruan}{Zhang
  et~al\mbox{.}}{2017}]%
        {zhang2017amulet}
\bibfield{author}{\bibinfo{person}{Pingping Zhang}, \bibinfo{person}{Dong
  Wang}, \bibinfo{person}{Huchuan Lu}, \bibinfo{person}{Hongyu Wang}, {and}
  \bibinfo{person}{Xiang Ruan}.} \bibinfo{year}{2017}\natexlab{}.
\newblock \showarticletitle{Amulet: Aggregating Multi-Level Convolutional
  Features for Salient Object Detection}. In
  \bibinfo{booktitle}{\emph{Proceedings of the IEEE Conference on Computer
  Vision and Pattern Recognition}}. \bibinfo{pages}{202--211}.
\newblock


\bibitem[\protect\citeauthoryear{Zhu and Xin}{Zhu and Xin}{2015}]%
        {zhu2015effective}
\bibfield{author}{\bibinfo{person}{Bing Zhu} {and} \bibinfo{person}{Yunhong
  Xin}.} \bibinfo{year}{2015}\natexlab{}.
\newblock \showarticletitle{Effective and robust infrared small target
  detection with the fusion of polydirectional first order derivative images
  under facet model}.
\newblock \bibinfo{journal}{\emph{Infrared Physics \& Technology}}
  \bibinfo{volume}{69} (\bibinfo{year}{2015}), \bibinfo{pages}{136--144}.
\newblock


\bibitem[\protect\citeauthoryear{Zhu, Cai, Kan, Thomas.H, and Ge}{Zhu
  et~al\mbox{.}}{2018a}]%
        {PDNet2018}
\bibfield{author}{\bibinfo{person}{Chunbiao Zhu}, \bibinfo{person}{Xing Cai},
  \bibinfo{person}{Huang Kan}, \bibinfo{person}{Li Thomas.H}, {and}
  \bibinfo{person}{Li Ge}.} \bibinfo{year}{2018}\natexlab{a}.
\newblock \showarticletitle{PDNet: Prior-model Guided Depth-enhanced Network
  for Salient Object Detection}. In \bibinfo{booktitle}{\emph{2018
  International Conference on Multimedia and Expo}}.
\newblock


\bibitem[\protect\citeauthoryear{Zhu, Huang, and Li}{Zhu
  et~al\mbox{.}}{2017a}]%
        {DBLP:journals/corr/abs-1710-04071}
\bibfield{author}{\bibinfo{person}{Chunbiao Zhu}, \bibinfo{person}{Kan Huang},
  {and} \bibinfo{person}{Ge Li}.} \bibinfo{year}{2017}\natexlab{a}.
\newblock \showarticletitle{Automatic Salient Object Detection for Panoramic
  Images Using Region Growing and Fixation Prediction Model}.
\newblock \bibinfo{journal}{\emph{CoRR}}  \bibinfo{volume}{abs/1710.04071}
  (\bibinfo{year}{2017}).
\newblock
\showeprint[arxiv]{1710.04071}
\urldef\tempurl%
\url{http://arxiv.org/abs/1710.04071}
\showURL{%
\tempurl}


\bibitem[\protect\citeauthoryear{Zhu, Huang, and Li}{Zhu
  et~al\mbox{.}}{2018b}]%
        {Zhu2018An}
\bibfield{author}{\bibinfo{person}{Chunbiao Zhu}, \bibinfo{person}{Kan Huang},
  {and} \bibinfo{person}{Ge Li}.} \bibinfo{year}{2018}\natexlab{b}.
\newblock \showarticletitle{An Innovative Saliency Guided ROI Selection Model
  for Panoramic Images Compression}. In \bibinfo{booktitle}{\emph{DCC}}.
  \bibinfo{pages}{438--438}.
\newblock


\bibitem[\protect\citeauthoryear{Zhu, Li, Guo, Wang, and Wang}{Zhu
  et~al\mbox{.}}{2017b}]%
        {Zhu2017A}
\bibfield{author}{\bibinfo{person}{Chunbiao Zhu}, \bibinfo{person}{Ge Li},
  \bibinfo{person}{Xiaoqiang Guo}, \bibinfo{person}{Wenmin Wang}, {and}
  \bibinfo{person}{Ronggang Wang}.} \bibinfo{year}{2017}\natexlab{b}.
\newblock \bibinfo{booktitle}{\emph{A Multilayer Backpropagation Saliency
  Detection Algorithm Based on Depth Mining}}.
\newblock \bibinfo{publisher}{Springer International Publishing},
  \bibinfo{address}{Cham}, \bibinfo{pages}{14--23}.
\newblock
\showISBNx{978-3-319-64698-5}
\urldef\tempurl%
\url{https://doi.org/10.1007/978-3-319-64698-5_2}
\showDOI{\tempurl}


\bibitem[\protect\citeauthoryear{Zhu, Li, Wang, and Wang}{Zhu
  et~al\mbox{.}}{2017c}]%
        {ZhuICCV2017}
\bibfield{author}{\bibinfo{person}{Chunbiao Zhu}, \bibinfo{person}{Ge Li},
  \bibinfo{person}{Wenmin Wang}, {and} \bibinfo{person}{Ronggang Wang}.}
  \bibinfo{year}{2017}\natexlab{c}.
\newblock \showarticletitle{An Innovative Salient Object Detection Using
  Center-Dark Channel Prior}. In \bibinfo{booktitle}{\emph{The IEEE
  International Conference on Computer Vision (ICCV)}}.
\newblock


\bibitem[\protect\citeauthoryear{Zhu, Li, Wang, and Wang}{Zhu
  et~al\mbox{.}}{2017d}]%
        {7966712}
\bibfield{author}{\bibinfo{person}{C. Zhu}, \bibinfo{person}{G. Li},
  \bibinfo{person}{W. Wang}, {and} \bibinfo{person}{R. Wang}.}
  \bibinfo{year}{2017}\natexlab{d}.
\newblock \showarticletitle{Salient Object Detection with Complex Scene Based
  on Cognitive Neuroscience}. In \bibinfo{booktitle}{\emph{2017 IEEE Third
  International Conference on Multimedia Big Data (BigMM)}}.
  \bibinfo{pages}{33--37}.
\newblock
\urldef\tempurl%
\url{https://doi.org/10.1109/BigMM.2017.22}
\showDOI{\tempurl}


\bibitem[\protect\citeauthoryear{Zhu, Li, Wang, and Wang}{Zhu
  et~al\mbox{.}}{2017e}]%
        {Zhu2017Salient}
\bibfield{author}{\bibinfo{person}{Chunbiao Zhu}, \bibinfo{person}{Ge Li},
  \bibinfo{person}{Wenmin Wang}, {and} \bibinfo{person}{Ronggang Wang}.}
  \bibinfo{year}{2017}\natexlab{e}.
\newblock \showarticletitle{Salient Object Detection with Complex Scene Based
  on Cognitive Neuroscience}. In \bibinfo{booktitle}{\emph{Multimedia Big Data
  (BigMM), 2017 IEEE Third International Conference on}}. IEEE,
  \bibinfo{pages}{33--37}.
\newblock


\end{thebibliography}

\end{document}